\documentclass[smallextended]{svjour3} 

\usepackage[most, breakable]{tcolorbox}  
\usepackage[utf8]{inputenc}  

\usepackage{xurl}
\usepackage{newtxmath} 
\usepackage{array}
\usepackage{multirow}
\usepackage{graphicx}
\usepackage{balance}
\usepackage{esvect}
\usepackage{booktabs}
\usepackage{tabularx}
\usepackage{anyfontsize}
\usepackage[all]{nowidow}
\usepackage[normalem]{ulem}
\usepackage{xcolor}
\usepackage{colortbl}
\usepackage{enumitem}
\usepackage{makecell}
\usepackage{caption}
\usepackage{subcaption}
\usepackage{framed}
\usepackage{wrapfig}

\usepackage{amsmath,amssymb}
\usepackage{natbib}
\usepackage{float}
\usepackage{lineno}
\usepackage{soul}
\usepackage{flushend}
\usepackage{multicol}
\usepackage{algorithm}
\usepackage{algpseudocode}
\usepackage{pifont}
\usepackage{rotating}
\usepackage{orcidlink}
\usepackage{adjustbox}

\definecolor{javagreen}{rgb}{0.25,0.5,0.35} 
\definecolor{gray}{rgb}{0.5,0.5,0.5}
\definecolor{lightgray}{rgb}{0.65,0.65,0.65}
\definecolor{codegreen}{RGB}{28,172,0}
\algnewcommand{\CommentG}[1]{\textcolor{codegreen}{// #1}}

\newcolumntype{C}[1]{>{\centering\arraybackslash}p{#1}}

\newcommand{\sectopic}[1]{\vspace*{0.1em}\par\noindent{\textit{\bfseries #1}}}

\renewcommand{\algorithmiccomment}[1]{\bgroup\color{gray}{//#1}\egroup}

\newcommand{\approach}{OPAL}

\let\cite\citep

\journalname{Empirical Software Engineering}

\newcounter{commentnumber}

\newcolumntype{Y}{>{\centering\arraybackslash}p{0.9cm}}

\newcommand{\rot}[1]{\rotatebox[origin=c]{60}{\makecell[l]{#1}}}

\begin{document}

\clearpage

\title{Effort-Optimized, Accuracy-Driven Labelling and Validation of Test Inputs for DL  Systems: A Mixed-Integer Linear Programming Approach} 
\titlerunning{Effort-Optimized, Accuracy-Driven Labelling and Validation of Test Inputs}
\author{Mohammad Hossein Amini, Mehrdad~Sabetzadeh, and Shiva~Nejati%
\thanks{M. H. Amini, M. Sabetzadeh, and S. Nejati are with the University of Ottawa, Canada.%
E-mail: \{mh.amini, m.sabetzadeh, snejati\}@uottawa.ca}}

\author{%
  Mohammad Hossein Amini\textsuperscript{1}\orcidlink{0009-0007-4312-7561}, 
  Mehrdad Sabetzadeh\textsuperscript{1}\orcidlink{0000-0002-4711-8319}, and 
  Shiva Nejati\textsuperscript{1}\orcidlink{0000-0002-0281-8231}%
}

\authorrunning{Amini et al.}

\institute{%
  Mohammad Hossein Amini \at \email{mh.amini@uottawa.ca}
  \and
  Mehrdad Sabetzadeh \at \email{m.sabetzadeh@uottawa.ca}
  \and
  Shiva Nejati \at \email{snejati@uottawa.ca} 
  \\
  \textsuperscript{1} University of Ottawa
}

\maketitle

\begin{abstract}
Software systems increasingly include AI components based on deep learning (DL). Reliable testing of such systems requires \emph{near-perfect} test-input validity and label accuracy, with \emph{minimal human effort}. Yet, the DL community has largely overlooked the need to build highly accurate datasets with minimal effort, since DL training is generally tolerant of labelling errors. This challenge, instead, reflects concerns more familiar to software engineering, where a central goal is to construct high-accuracy test inputs -- with accuracy as close to 100\% as possible -- while keeping associated costs in check. In this article, we introduce \approach, a human-assisted labelling method that can be configured to target a desired accuracy level while minimizing the manual effort required for labelling. The main contribution of \approach\ is a mixed-integer linear programming (MILP) formulation that minimizes labelling effort subject to a specified accuracy target. To evaluate \approach, we instantiate it for two tasks in the context of testing vision systems: automatic labelling of test inputs and automated validation of test inputs. Our evaluation, based on more than 2500  experiments performed on nine datasets, comparing \approach\ with eight baseline methods,  shows that \approach, relying on its MILP formulation, achieves an average  accuracy of 98.8\%, just 1.2\% below perfect accuracy, while cutting  manual labelling by more than half. Further, \approach\ significantly outperforms automated labelling baselines in labelling accuracy across all nine datasets, with large effect sizes, when all methods are provided with the same manual-labelling budget. For automated test-input validation, on average,  \approach\ reduces manual effort by 28.8\% while achieving 4.5\% higher accuracy than the  state-of-the-art test-input validation baselines. Finally, we show that augmenting \approach\ with an active-learning loop leads to an additional 4.5\% reduction in required manual labelling, without compromising accuracy.
\end{abstract}


\section{Introduction}
\label{sec:intro}

Deep learning (DL) has become a cornerstone for numerous advanced applications, enabling the automation of complex tasks such as anomaly detection, object recognition, and semantic segmentation. Since DL is inherently data-driven, accurately testing and verifying DL systems  necessitates large and diverse test sets. To address this need, researchers have developed a variety of synthetic test-input generation methods~\cite{Pei_2019_Deepxplore, Dola_2021_DAIV, marsha, WhenWhyTest2022, Tian_2018_deeptest, Zhang2018Deeproad}. For example, a common approach for testing vision-based  systems is to generate synthetic images, either by perturbing existing images~\cite{WhenWhyTest2022, Pei_2019_Deepxplore, Guo_2018_DLFuzz} or by using generative models~\cite{Dola_2021_DAIV, Zhang2018Deeproad}. 
However, recent research indicates that these methods frequently produce images where the original or intended labels are not preserved, requiring manual review and relabelling~\cite{marsha, WhenWhyTest2022,Ghobari_2025_active_learning}. It is therefore important to find ways to reduce the effort required for manual labelling.

Both supervised and semi-supervised learning have been explored to reduce the effort required for labelling large volumes of data, such as synthetic or web-crawled datasets~\cite{Pham_2021_MetaPseudoLabel, Chen_2020_Contrastive, Lee_2013_PseudoLabel}. Existing methods developed for this purpose typically use pretrained DL classifiers to generate provisional labels for unlabelled data. This provisionally labelled data is then merged -- either fully or selectively -- with the manually labelled data, usually for re-training or fine-tuning~\cite{Pham_2021_MetaPseudoLabel, Lee_2013_PseudoLabel}. This process can considerably decrease reliance on manual labelling. Nevertheless,  existing methods pay little attention to the accuracy of the automatically generated labels. This is because DL training algorithms often tolerate moderate levels of labelling errors -- around 10\%, according to existing studies~\cite{Duan_2023_SufficientLabels, Song_2023_LabelNoise} -- without major compromises in accuracy, particularly when the training sets are large. 

In contrast, when it comes to testing DL systems, it is essential that the test inputs used be highly accurate to ensure reliable testing results~\cite{DBLP:journals/computer/AdigunCFGMPRS22}. Even a small number of mislabelled elements in a test set can lead to either inflated or deflated measured accuracy of the system under test. Inflated accuracy can create a false sense of confidence in the system's trustworthiness, making it more likely that genuine faults will be overlooked, potentially leading to critical failures. Deflated accuracy, on the other hand, misrepresents trustworthiness by generating spurious errors, i.e., false issues caused by label inaccuracies rather than actual system flaws. Such errors waste time and effort by prompting investigations into non-existent issues and can also potentially divert attention from identifying and resolving genuine faults.

Reliable testing requires \emph{near-perfect} test-set accuracy and \emph{minimal human effort}. Yet, the DL community has largely overlooked this issue, since, as stated above, DL training is generally tolerant of labelling errors. \textbf{The challenge, instead, reflects concerns more familiar to software engineering, where a central goal is to construct high-accuracy test inputs while keeping associated costs in check.}

We therefore argue that, for software verification purposes, data labelling should be approached as an \emph{accuracy-driven} process, in which explicit control is provided over the desired accuracy threshold, and effort is optimized subject to this threshold: for test inputs, the goal would typically be to achieve near-perfect label accuracy, and effort needs to be optimized subject to this goal, whereas for training DL systems, effort should be optimized according to a more relaxed criterion -- namely, ensuring that the prevalence of mislabelling remains within the tolerance margin of the training algorithms.

In this article, we introduce \approach, an effort-\textbf{OP}timized  \textbf{A}ccuracy-driven  data \textbf{L}abelling method. \approach\ aims  to meet or exceed a specified accuracy target with the least possible human effort. To do so, \approach\ identifies difficult cases that classifiers cannot confidently label, directing them to humans for handling, while routine cases are automatically labelled by classifiers. \approach\ relies on the predicted labels and prediction confidences obtained from a set of  pretrained classifiers. It then uses an optimization solver to determine when the classifiers' confidence levels are  high enough to bypass manual labelling. The main novelty of \approach\ is to formulate the optimization problem of minimizing human labelling effort under a specified accuracy threshold as a \emph{mixed-integer linear program (MILP)}~\cite{Boyd2014Convex, Thie2008LinearProgramming} and to solve it efficiently.

Given the critical importance of validity and accuracy of test inputs in software testing, we evaluate \approach\ for two tasks pertinent to testing vision-based systems: (1) automated test-input labelling, and (2) automated test-input validation~\cite{marsha,WhenWhyTest2022,Ghobari_2025_active_learning}. 
Automated test-input labelling is about generating accurate labels for individual test images, whereas, following the definition of Riccio and Tonella~\cite{WhenWhyTest2022}, automated test-input validation is about determining whether a transformed image remains within the domain of its original image and can be confidently labelled by a human domain expert.

We apply \approach\ to nine image datasets: six real-world  benchmarks and three synthetic datasets. We instantiate \approach\ using a set  of three widely used DL classifiers: VGG~\cite{vgg}, ResNet~\cite{resnet}, and ViT~\cite{vit}. These represent complementary state-of-the-art DL architectures for vision-based systems: VGG is a convolutional neural network (CNN), ResNet extends CNNs with residual connections, and ViT employs transformers. In our experiments, we use versions of these classifiers pretrained on the ImageNet~\cite{imagenet} dataset, which is not part of any of our nine evaluation datasets.

We compare \approach\ with eight baselines. Four are state-of-the-art supervised or semi-supervised automated data-labelling techniques~\cite{Lee_2013_PseudoLabel} from the ML and DL literature implemented with VGG, ResNet, and ViT backbones~\cite{vgg,resnet,vit}.
The other four are automated data-validation approaches taken from the software engineering literature, specifically from software testing~\cite{marsha,WhenWhyTest2022,Ghobari_2025_active_learning}. Our evaluation relies on standard metrics: labelling accuracy and the amount of manual  effort required.
To assess \approach\ for data labelling we use all nine of our datasets; to assess it for data validation we use the three datasets from the software engineering literature that contain both the original images and their transformed variants, making them suitable for the data-validation use case.

Our results obtained from more than 2500 experiments  show that, across nine diverse
datasets, when \approach\ is configured to target 100\% accuracy, it achieves an average accuracy of 98.8\%, just 1.2\% below perfect accuracy, while requiring 44.2\% manual effort, effectively cutting manual labelling by more than half. Further, \approach\  significantly outperforms the supervised and semi-supervised baselines in labelling accuracy for all nine datasets with large effect sizes. Under the same manual labelling effort required by \approach\  to achieve an average accuracy of 98.8\%, none of the baselines exceed an average accuracy of 94.5\%.   For automated test-input validation, on average,  \approach\ reduces manual effort by 28.8\% while achieving 4.5\% higher accuracy than the four state-of-the-art test-input validation baselines~\cite{marsha,WhenWhyTest2022,Ghobari_2025_active_learning}. Statistical tests confirm that both the reduction in manual effort and the improvement in accuracy are statistically significant, with a large effect size for manual effort reduction across all datasets and medium to large effect sizes for accuracy improvement across all datasets. We further show that incorporating an active-learning loop into \approach\ reduces the required manual labelling by an additional 4.5\% without compromising accuracy. Finally, \approach\ has practical execution time. Specifically, our results indicate that \approach's main contribution -- its MILP-based optimization step -- is highly efficient, averaging only 24 seconds per dataset. This accounts for approximately 3\% of the total execution time and is the only additional computational overhead compared to baselines that do not use accuracy-driven effort minimization.

\section{\approach\ at a Glance}
\label{sec:motivate}
This section outlines  \approach\ and its novelty and significance. 

\textbf{Overview.} Given an unlabelled dataset \(D\) and a set of pretrained classifiers, \approach\ assigns a label to each element \(e \in D\) automatically if all classifiers agree on the same label with high confidence; otherwise, \(e\) is labelled manually. The ultimate goal of \approach, expressed as a constrained optimization problem below, is to \emph{minimize the amount of manual labelling required while ensuring that the labelling accuracy meets or exceeds a threshold \(\alpha\):} 

\vspace*{-.5cm}
\begin{align}
    & \text{minimize}
\qquad \text{Manual Effort} \tag{\ding{81}}\label{eqn:objective} \\
& \text{subject to}
\qquad \text{Accuracy} \geq \alpha \notag 
\end{align}
\vspace*{-.5cm}

The above constrained optimization problem computes the decision variables that determine when the classifiers' confidence levels are sufficiently high to bypass manual labelling. However, computing these decision variables for a given dataset requires access to ground-truth labels for the dataset. Without ground-truth labels, one cannot measure the accuracy of the classifiers' predictions and cannot ensure the $\text{Accuracy} \geq \alpha$ constraint.

Hence, \approach\ selects a small subset \(D_o \subset D\), referred to as the \emph{optimization} subset, and obtains its ground-truth labels through manual labelling.  It then solves problem~(\ref{eqn:objective}) on \(D_o\) to determine the classifiers’ confidence levels that minimize manual labelling while ensuring classification accuracy meets or exceeds \(\alpha\) for $D_o$. Next, \approach\ applies these classifiers’ confidence levels  to label the remaining unlabelled elements in \(D\). Specifically, when all classifiers agree on a label for an unlabelled element $e \in D$ and their confidence meets or exceeds the levels derived from solving problem~(\ref{eqn:objective}), $e$ is labelled automatically; otherwise, it is labelled manually.

Because \approach\ solves problem~(\ref{eqn:objective}) only for the manually labelled subset \(D_o\), it cannot guarantee an optimal solution for the entire dataset \(D\). Nonetheless, \approach\ uses three mechanisms 
to ensure that its labelling performance across $D$ closely approximates an optimal solution to problem~(\ref{eqn:objective}).  First, before solving problem~(\ref{eqn:objective}), \approach\ fine-tunes the underlying classifiers using a separate manually labelled subset \(D_t \subset D\) -- disjoint from \(D_o\) -- referred to as the \emph{fine-tuning} subset, to ensure high classification accuracy on \(D\). Second, \approach\ uses diversified random sampling to select   \(D_t\) and \(D_o\), ensuring they are representative of $D$. Finally, the decision variables derived from exhaustively solving problem~(\ref{eqn:objective}) on \(D_o\) are applied to the entire dataset \(D\). Our empirical evaluation in Section~\ref{sec:eval} shows that these mechanisms enable \approach\ to achieve near-optimal -- if not exactly optimal -- solutions on the full dataset and that \approach\  significantly outperforms labelling methods that do not  use constrained optimization.

\textbf{Novelty.} The key novelty of \approach\ is its accuracy-driven labelling process, which explicitly controls a target accuracy threshold while optimizing labelling effort. Further, we show that \approach's labelling process -- corresponding to problem~(\ref{eqn:objective}) -- can be formulated as an MILP and solved efficiently. To our knowledge, our effort-optimized, accuracy-driven formulation for data labelling is novel. Although one could, in principle, solve problem~(\ref{eqn:objective}) using alternatives such as genetic algorithms~\cite{metaheuristicsbook} or a brute-force search, these approaches have notable limitations. Heuristic methods (e.g., genetic algorithms)  do not provide exhaustive search or deterministic reproducibility, whereas MILP solvers~\cite{Boyd2014Convex,Thie2008LinearProgramming,Mitchell_2011_pulp} ensure both.  Brute-force approaches are also impractical due to their inability to handle floating-point decision variables and their poor scalability with increasing dimensionality -- our formulation involves real-valued decision variables and the number of constraints in our formulation  grows with the number of classifiers. Finally, while machine learning models can be used for optimization tasks, they cannot solve constrained optimization problems that require strict enforcement of constraints (e.g., ensuring \(\text{Accuracy} \geq \alpha\)). Consequently, MILP remains the preferred approach for our formulation.

\textbf{Significance.} A 2021 study by Northcutt et al.~\cite{Curtis_2021_TestLabel} found that many labels in the test sets of popular datasets are incorrect, for example, reporting labelling error rates of 5.8\% for ImageNet~\cite{imagenet} and 10.1\% for QuickDraw~\cite{quickdraw}. The authors observed that mislabelled test sets can significantly distort model validation and verification. Notably, in their experiments, models that performed poorly on  benchmarks such as ImageNet and QuickDraw turned out to be among the most accurate once the labelling errors in the benchmarks were corrected~\cite{Curtis_2021_TestLabel}. \approach\ aims to improve the quality of data labelling, particularly in the context of testing, by enabling practitioners to directly control their accuracy target while minimizing labelling effort accordingly. To achieve this, \approach\ identifies difficult instances that, based on the chosen accuracy target, pretrained classifiers cannot confidently label. These instances are delegated to human annotators, while the remaining data is labelled automatically.

\section{Background}

\subsection{Mixed-Integer Linear Programming (MILP)}
Constrained optimization finds values for decision variables that satisfy a set of equality and inequality constraints while minimizing or maximizing an objective function~\cite{Boyd2014Convex}. Linear Programming (LP) is a subclass of constrained optimization techniques in which both the objective and the constraints are linear and all variables are continuous. Mixed-Integer Linear Programming (MILP) extends LP by enforcing integrality on some variables to capture discrete decisions (e.g., selecting modules or features).
In software engineering, constrained optimization techniques such as MILP and LP have been applied to reliability optimization, effort estimation, decision support, and product-line configuration, e.g., \cite{BermanAshrafi1993,Ruhe2020,HenardPapadakisHarmanLeTraon2015}. In this article, we formulate the tasks of test-input labelling and test-input validation as MILP problems, with the goal of building high-accuracy test sets while minimizing manual effort.

\subsection{Testing Vision Systems}
\label{subsec:testing_vision_systems}
As motivated in Section~\ref{sec:intro}, we evaluate \approach\ by applying it in the context of vision-system testing. 
Vision systems, being based primarily on deep learning and machine learning, do not have traditional specifications~\cite{Dola_2021_DAIV}. This leads to two important challenges in testing vision systems: (Challenge 1) Developing test oracles, i.e., test-input labels, remains largely a manual task; and (Challenge 2) Test-generation methods for such systems, e.g., metamorphic testing~\cite{marsha,WhenWhyTest2022,Guo_2018_DLFuzz,DBLP:conf/aitest/ArcainiBBG20}, may produce invalid test inputs, necessitating further automated measures to  determine test-input validity.

\begin{figure}[t]
    \centering
    \includegraphics[width=0.8\linewidth]{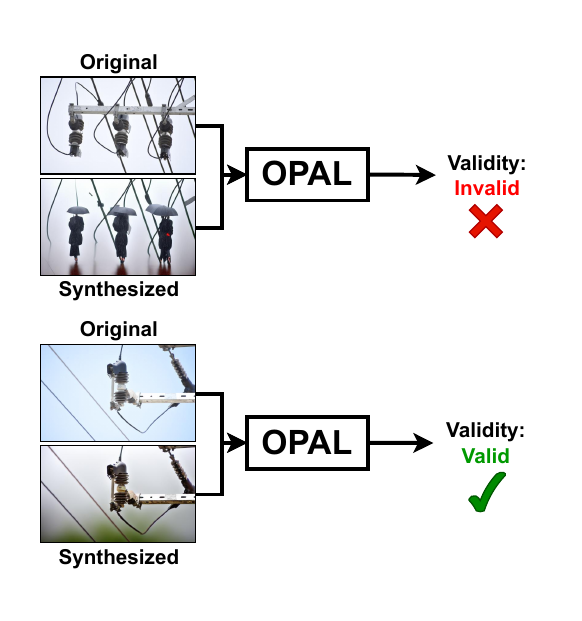}
\vspace*{-.6cm}
    \caption{Illustrating the application of \approach\ for test-input validation. An image and its transformation are provided to \approach. \approach\ determines whether the transformed image is a ``valid'' alteration of the original image. }
    \label{fig:opal_example}
\end{figure}

\approach\ helps address both of the above challenges. With regard to the first challenge,  \approach\ can be employed to automatically produce labels, i.e., test oracles, for test data, providing explicit control over the trade-off between accuracy and manual effort. We refer to such a use case of \approach\ as \textbf{automated test-data labelling}. With regard to the second challenge, when provided with a pair of test-data images -- where one is the original image and the other is a transformed version of the original -- \approach\ can determine whether the pair is valid or not. 
For instance, Figure~\ref{fig:opal_example} 
shows an example where a generative AI technique  -- more specifically, Stable Diffusion~\cite{brooksInstructPix2PixLearningFollow2023}-- has added a rain effect to an original image. In this example, Stable Diffusion is prompted with an original image and a textual description to transform the image so that it includes the intended climatic conditions (rain, fog, or snow) at different magnitude levels (light, moderate, and intense).
The task is to decide whether the resulting image is ``valid'' or ``invalid''.  
In this example, following the validity definition of Riccio and Tonella~\cite{WhenWhyTest2022}, a ``valid'' image is one that is within the real-world image domain, i.e., confidently recognized as a utility pole by a domain expert, whereas an ``invalid'' image either does not represent a utility pole or is too distorted to be confidently identified as one.
Here, \approach\ identifies the top transformation in Figure~\ref{fig:opal_example} as an invalid test input, since in the transformation, the high-voltage fuse cutouts on the utility pole have been mistaken as humans, and the rain effect has been added by altering the cutouts and giving them umbrellas. The bottom pair is identified as valid, since the transformed image retains the content of the original image and only modifies the weather condition in the original image. We refer to this second use case of \approach\ as \textbf{automated test-data validation}. As with the first use case, \approach\ provides explicit control over the trade-off between accuracy and manual effort.

In this article, we systematically evaluate \approach\ for both the automated test-data labelling use case (RQ1 and RQ2 in Section~\ref{sec:eval}) and the automated test-input validation use case (RQ3 in Section~\ref{sec:eval}). Our evaluation results show that \approach\ is an effective tool for both use cases in the context of testing vision systems, outperforming state-of-the-art baselines.

\subsection{Active Learning}
\label{subsec:active_learning}
Active learning is an iterative, human-in-the-loop approach designed to improve the accuracy of machine learning models while reducing annotation effort. In each iteration of active learning, only the most uncertain samples are sent to human annotators for labelling, while the rest are automatically labelled by the model. The model is then retrained on the expanded labelled dataset that includes the newly annotated samples~\cite{Settles2009}.
This approach concentrates human effort where it most benefits accuracy, thus lowering manual effort~\cite{Settles2009, Ghobari_2025_active_learning}. 
Recent work has shown that active learning can be effectively applied to test-input validation by training lightweight machine-learning models on image features~\cite{Ghobari_2025_active_learning}. Therefore, to evaluate \approach\ for test-input validation, we also consider an extension with active learning to examine its potential for further reducing manual effort.

\section{Optimized Labelling via MILP}
\label{sec:problem}
Algorithm~\ref{alg:optimzationlabel} details \approach.  It  takes as input an unlabelled dataset $D$,  an accuracy threshold $\alpha$, a set of $n$ pretrained classifiers, a percentage $\mathit{hInitial}$ of $D$ to be manually labelled initially, and the size $s$ of the subset $D_o \subset D$ used for optimization. Ultimately, for our evaluation,  we measure  the proportion of \(D\) labelled manually versus automatically. Hence, we define  \emph{hInitial} as the fraction of  \(D\) that \approach\  requires to be manually labelled initially (i.e., the set \(D_t \cup D_o\)), and then partition this fraction into \(D_t\) and \(D_o\) so that \(|D_o| = s\).   We specify \(|D_o|\) directly via the input parameter \(s\), rather than as a fraction of \(D\), because \(|D_o|\) determines the number of constraints in the MILP used to solve problem~\eqref{eqn:objective}. The output of \approach\ is the labelled dataset $D$, with some elements manually labelled and the rest labelled by the classifiers.

\footnotesize{
\begin{algorithm}[t]
\caption{Effort-Optimized Accuracy-Driven Labelling (\approach) using mixed-integer linear programming (MILP)}
\label{alg:optimzationlabel}

\begin{flushleft}
\textbf{Input} $\mathit{D}$: An unlabelled dataset \\ 
\textbf{Input} $\alpha$: Accuracy threshold \\
\textbf{Input} $c_1, \ldots, c_n$: A set of pretrained classifiers\\
\textbf{Param} $\mathit{hInitial}$:  Percentage of $D$  for fine-tuning and decision variable computation \\
\textbf{Param} $s$: Number of elements used for computing decision variables using MILP\\[.5em]
\textbf{Output} $D$: Labelled dataset\\
\end{flushleft}
\begin{algorithmic}[1]
\Statex \CommentG{\textbf{Step 1. Data splitting}} 
\State $(D_t,  D_o, D')=\text{DataSplit}(D, \mathit{hInitial}, s)$ \CommentG{$\frac{|D_t \cup D_o|}{|D|} = \mathit{hInitial}, |D_o|= s$} 
\Statex \CommentG{\(D_t, D_o,\) and \(D'\) are for fine-tuning, optimization, and labelling, respectively.}
\State Manually label $D_t \cup D_o$
\Statex \CommentG{\textbf{Step 2: Fine tuning}}
\State $\text{FineTune}\big((c_1, \ldots, c_n), D_t)\big)$  
\Statex \CommentG{\textbf{Step 3: Computing decision variables using MILP}}
\State $(\omega_1, \ldots, \omega_n)= \text{MILPSolver}\big((c_1, \ldots, c_n), D_o, \alpha \big)$  
\Statex \CommentG{\textbf{Step 4: Labelling $D'$ using $\omega_1, \ldots, \omega_n$}}
\State $D_m = \varnothing$   \CommentG{ A subset of $D'$ to be manually labelled on Line 17}
\For{$\mathit{e}_i \in D'$} 
    \For{$j \in \{1..n\}$}
    \State Let $\hat{l}_{ji}$ be the predicted label for $\mathit{e}_i$ by $c_j$ 
     \State Let $\hat{\theta}_{ji}$ be the prediction confidence for $\mathit{e}_i$ by $c_j$ 
     \EndFor
\Statex \CommentG{Labelling decision (\textbf{Condition~1} and \textbf{Condition~2})}    
    \If{$\hat{l}_{1i} = \dots = \hat{l}_{ni}$ and $\omega_1\hat{\theta}_{1i}$ + \dots + $\omega_n\hat{\theta}_{ni} > 1$}
    \State Label($\mathit{e}_i$) = $\hat{l}_{1i}$  
    \Else
    \State $D_m = D_m \cup \{e_i\}$
    \EndIf
\EndFor
\State Manually label $D_m$
\State \Return  the labelled set $D$
\end{algorithmic}
\end{algorithm}
}
\normalsize

\textbf{Assumptions.}  \approach\ makes two assumptions about the classifiers \(c_1, \dots, c_n\): (1) they are pretrained, and (2) for each element \(e \in D\), each classifier labels $e$ with a  confidence indicating how certain the classifier is about its prediction. 
While the formulation of \approach\ does not inherently restrict the type or nature of data in $D$, in this article, we evaluate \approach\ exclusively using image data. Extending the evaluation of \approach\ to other data modalities, such as text, is left for future work.

The four steps of Algorithm~\ref{alg:optimzationlabel} are  discussed below:

\textbf{Data Splitting (Step~1).}  \approach\  partitions the unlabelled dataset $D$ into three  subsets: a \emph{fine-tuning subset} $D_t$, an \emph{optimization subset}  $D_o$, and a \emph{to-be-labelled subset} denoted by $D'$ (Line~1). The subset $D_t$ is used to fine-tune the pretrained classifiers in Step~2;  $D_o$ is used for optimization in Step~3; and in Step~4, each element in $D'$ is labelled either automatically (Line~12) or manually (Line~17).

Specifically, \approach\ partitions \(D\) into clusters by grouping elements with high similarity (or proximity) according to a relevant distance metric. Next, it draws an equal number of elements from each cluster until an \(\mathit{hInitial}\) proportion of \(D\) is selected. This sampling strategy increases the diversity of the selected samples and helps them be more representative of \(D\).  The sampled subset is then  divided randomly into two sets: $D_t$ and $D_o$ such that \(|D_o| = s\). The sets $D_t$ and $D_o$ are then manually labelled  (Line~2). 

\textbf{Fine-tuning (Step~2).}   \approach\ fine-tunes the classifiers  using the subset $D_t$  (Line~3). Since retraining from scratch is both computationally expensive and requires extensive labelled data, as mentioned earlier in our assumptions, \approach\ leverages pretrained classifiers. Fine-tuning the classifiers on the subset \(D_t\) increases their accuracy for the dataset \(D\). 

\textbf{Computing decision variables using MILP (Step~3).}
\approach\ formulates  the constrained optimization problem~(\ref{eqn:objective}) as an MILP problem and solves it for the set $D_o$  (Line~4). For each $\mathit{e}_i \in D_o$, each classifier $c_{j}$ generates a label $\hat{l}_{ji}$ with confidence  $\hat{\theta}_{ji}$.  
We introduce decision variables \(\omega_1, \ldots, \omega_n\) corresponding to classifiers $c_1, \ldots, c_n$. Each variable $\omega_j$ is a weight signifying how much credibility should be given to classifier $c_j$ for labelling. 
We note that the weights assigned to classifiers are not fixed or manually specified. Instead, they are decision variables optimized by the MILP. The motivation for allowing different weights is that, following fine-tuning, classifiers may differ in classification accuracy, and thus in the reliability of their predictions. 
Allowing non-uniform weights gives the solver the flexibility to rely more on accurate classifiers and less on weaker ones when doing so reduces human effort while still satisfying the accuracy constraint.
\approach\ automatically labels an element $\mathit{e}_i$ when these two conditions hold: 

\begin{description}
 \item[(Condition~1)] $\hat{l}_{1i} = \dots = \hat{l}_{ni}$ (i.e., all classifiers label \( e_i \) the same) 
 \item[(Condition~2)] The weighted sum  $\omega_1\hat{\theta}_{1i} + \dots + \omega_n\hat{\theta}_{ni}$ is higher than a threshold 
\end{description}
Without loss of generality, we assume the threshold is 1 (one), as any other value would allow us to normalize all the weights, \( \omega_1, \ldots, \omega_n \), by dividing them by that value.  Specifically, we write Condition~2 as follows: 
\hbox{$\omega_1\hat{\theta}_{1i} + \dots + \omega_n\hat{\theta}_{ni} > 1$}.

For each \(\mathit{e}_i \in D_o\), we define the predicate \(z_i\) and the variable \(x_i\) to represent Condition~1 and Condition~2 for \(e_i\), respectively:

\vspace*{-.3cm}
\begin{equation}
    z_i = \begin{cases}
        1; & \hat{l}_{1,i} = \dots = \hat{l}_{n,i} \\
        0; & \text{Otherwise}
    \end{cases},  \quad
    x_i = \begin{cases}
        1; & \omega_1\hat{\theta}_{1i} + \dots + \omega_n\hat{\theta}_{ni} > 1 \\
        0; & \text{Otherwise}
    \end{cases} \label{eqn:x_i}
\end{equation}

Predicates $z_i$ have known values. However, variables $x_i$  depend on the weights $\omega_{j}$ that need to be determined via optimization. Hence, similarly to the weights \(\omega_j\), the variables \(x_i\) are also considered decision variables in our formulation.

We formalize the Manual Effort in  problem (\ref{eqn:objective}) as the number of elements that would be assigned to the human labeller, noting that  \approach\  automatically labels $\mathit{e}_i$ if Conditions~1 and~2 hold --  i.e.,  both $x_i$ and $z_i$ are one. Otherwise,  $\mathit{e}_i$ 
is assigned to a human labeller if $x_iz_i=0$.

\vspace*{-.5cm}
\begin{align}
    \text{Manual Effort} &= |\{\mathit{e}_i \in D_o \mid z_ix_i = 0\}| = \sum_{i=1}^{|D_o|}(1 - z_ix_i) \label{eqn:final_objective}
\end{align}
\vspace*{-.3cm}

We now formalize Accuracy in  problem (\ref{eqn:objective}). Accuracy is defined as the proportion of elements in \(D_o\) that are labelled correctly by \approach, whether automatically or manually. Only those elements labelled automatically by \approach\ can potentially be labelled incorrectly. We define the intermediary predicates \(b_i\) to indicate whether all classifiers label \(\mathit{e}_i\) correctly according to its ground-truth label:

\vspace*{.05cm}
\begin{center}
$\begin{array}{lcl}
     b_i &= & \begin{cases}
         1; \quad l_i = \hat{l}_{1,i} = \dots = \hat{l}_{n,i}\\
         0; \quad \text{Otherwise}
     \end{cases}
\end{array}$
\end{center}
\vspace*{.05cm}

where  $l_i$ is the ground-truth label of  $\mathit{e}_i$. \approach\  labels an element \(\mathit{e}_i\) incorrectly (and hence automatically) if \(x_i z_i = 1\) and \(x_i b_i = 0\). Stated otherwise, \approach\ correctly labels $e_i$ if \(x_i z_i = 0\) or \(x_i b_i = 1\). Given this, we formalize accuracy as the complement of the proportion of elements in \(D_o\) that are incorrectly (and automatically) labelled.

{\fontsize{8.3}{7}\selectfont
\begin{align}
    \text{Accuracy}  &= 1 - \frac{\sum_{i=1}^{|D_o|} (z_ix_i - b_ix_i)  }{|D_o|} = \frac{|D_o| + \sum_{i=1}^{|D_o|} (b_i - z_i)x_i }{|D_o|} \notag
\end{align}}

We now rewrite the constraint  \(\text{Accuracy} \geq \alpha\) as follows: 

{\fontsize{8}{7}\selectfont
\begin{align}
    \frac{|D_o| + \sum_{i=1}^{|D_o|} (b_i - z_i)x_i}{|D_o|} &  \geq \alpha \Longrightarrow \sum_{i=1}^{|D_o|} (b_i - z_i)x_i + |D_o|(1 - \alpha)\geq 0     \label{eqn:final_accuracy} 
\end{align}}

Equation~\ref{eqn:x_i} relates variables $x_i$ to  weights $\omega_j$ through a case-based function which is non-linear. Following the standard practice in linear programming~\cite{Thie2008LinearProgramming, Boyd2014Convex},  we linearize this relationship by introducing the following two constraints:

\vspace*{-.3cm}
\begin{align*}
    \omega_1\hat{\theta}_{1i} + \dots + \omega_n\hat{\theta}_{ni} - 1 &\leq M x_i\\
    \omega_1\hat{\theta}_{1i} + \dots + \omega_n\hat{\theta}_{ni} - 1 &\geq -M(1 - x_i) + \epsilon
\end{align*}
\vspace*{-.3cm}

where $M$ is a large constant, e.g., $10^6$, and $\epsilon$ is a small positive value, e.g., $10^{-6}$. These constraints force each variable $x_i$ to be either one when $\omega_1\hat{\theta}_{1i} + \dots + \omega_n\hat{\theta}_{ni} > 1$, or  zero when \hbox{$\omega_1\hat{\theta}_{1i} + \dots + \omega_n\hat{\theta}_{ni} \leq 1$}. Note that when \hbox{$\omega_1\hat{\theta}_{1i} + \dots + \omega_n\hat{\theta}_{ni} = 1$}, the presence of $\epsilon$ in the second inequality ensures that $x_i$ is 0.

Using Equation~(\ref{eqn:final_objective}), Constraint~(\ref{eqn:final_accuracy}) and the above two constraints, we obtain the following characterization of our constrained optimization problem~(\ref{eqn:objective}):

\resizebox{0.97\columnwidth}{!}{%
  \begin{minipage}{\columnwidth}
\begin{align}
\underset{\omega_1, \dots, \omega_n,x_1, \dots, x_{|D_o|}}{\text{minimize}} \quad & |D_o| - \sum_{i=1}^{|D_o|} z_ix_i & & \nonumber \\
\text{subject to} \quad & \sum_{i=1}^{|D_o|} (b_i - z_i)x_i + |D_o|(1 - \alpha) & \geq & \ 0  & \wedge \notag \\
& \bigwedge_{ i = 1, \dots, |D_o|} \omega_1 \hat{\theta}_{1i} + \cdots + \omega_n \hat{\theta}_{ni} - 1 & \leq & \  M x_i & \wedge  \notag \\
& \bigwedge_{i = 1, \dots, |D_o|} \omega_1 \hat{\theta}_{1i} + \cdots + \omega_n \hat{\theta}_{ni} - 1 & \geq & \ -M(1 - x_i) + \epsilon  &\notag
\end{align}
\end{minipage}
}

\hspace*{0.5cm}

The minimization objective and the constraints in the above are all linear with respect to the floating-point decision variables $\omega_1, \dots, \omega_n$,  and the binary decision variables $x_1, \dots, x_{|D_o|}$. 
Hence,  the above optimization problem is an MILP problem and can be solved using MILP solvers~\cite{Boyd2014Convex, Thie2008LinearProgramming, Mitchell_2011_pulp}.  Our MILP formulation includes  $n + |D_o|$ decision variables and $2|D_o| + 1$ constraints.  As demonstrated in Section~\ref{sec:exp-res}, the MILP solver takes significantly less 
time
than the other three steps  while enabling \approach\ to achieve a very high level of accuracy.

\textbf{Labelling (Step~4).} \approach\ labels $D'$, i.e., the remaining unlabelled subset of  $D$, using the  $\omega_1, \ldots, \omega_n$ values obtained from solving the constrained optimization problem in Step~3. First, each classifier generates a label and a confidence score for each element in $D'$  (Lines~7--10). Then, if both Conditions~1 and~2 hold, elements are automatically labelled. Otherwise, elements are stored in a set $D_m$ to be manually labelled at the end (Line~17). Once all the elements  are labelled, \approach\ returns the labelled set $D$ (Line~18).

\section{Evaluation Setup}
\label{sec:eval}
We evaluate \approach\ on nine image datasets: six containing real-world images and three containing synthetic images specifically generated for testing vision-based systems. Our evaluation aims to answer the three research questions (RQs) described below. As discussed in Section~\ref{subsec:testing_vision_systems}, \approach\ can be used in two distinct scenarios for improving the testing of DL systems: (1) using \approach\ for test-data labelling, and (2) using \approach\ for test-input validation. RQ1 and RQ2 evaluate \approach\ for test-data labelling and compare it against test-data labelling baselines, whereas RQ3 evaluates \approach\ for test-input validation and compare it against test-input validation baselines.

\textbf{RQ1. (OPAL's test-data labelling effort and accuracy)} \emph{How close does \approach\ come to perfect accuracy, and how much manual labelling is required to achieve peak (near-perfect) accuracy?} We configure \approach\  by setting \(\alpha\) in Algorithm~\ref{alg:optimzationlabel} to \(100\%\) to push \approach\ to reach its upper bound in terms of accuracy. We then evaluate \approach\ by systematically varying the algorithm's inputs, including the set of input classifiers and the parameter \emph{hInitial}, which represents the amount of manually labelled data used for fine-tuning and optimization. 

We measure both the labelling accuracy and the total labelling effort. Note that the parameter \emph{hInitial} determines the initial labelled set for fine-tuning and optimization, while the total effort additionally includes the set \(D_m\), which is manually labelled at the end of Algorithm~\ref{alg:optimzationlabel}. Based on these results, we provide insights into how many classifiers and how much pre-labelled data (\emph{hInitial}) would be required for \approach\ to achieve near-perfect accuracy with minimal overall labelling effort.

\textbf{RQ2. (Comparison with state-of-the-art test-data labelling baselines)}
\emph{Given the same manual labelling effort used by \approach\ in RQ1 to achieve near-perfect accuracy, can state-of-the-art methods achieve a comparable level of labelling accuracy?}  RQ2 compares \approach\ with state-of-the-art supervised and semi-supervised methods for image labelling. We provide the baselines with the same amount of manually labelled elements  that \approach\ needs for reaching near-perfect accuracy in RQ1. We then compare their labelling accuracy to \approach's.  

\textbf{RQ3. (OPAL for test-input validation)} \emph{How does OPAL compare to state-of-the-art baselines for test-input validation?}
In this research question, we evaluate \approach\ for test-input validation where \approach\ determines whether a transformed image (test input) is a valid alteration of its corresponding original image or not. As discussed in Section~\ref{sec:intro}, a synthesized test input is considered valid if it remains within the domain of its original counterpart, i.e., can be confidently labelled by a domain expert~\cite{WhenWhyTest2022}. We conduct the evaluation in two ways.
First, we apply \approach\ as described in Algorithm~\ref{alg:optimzationlabel} for validating test inputs. Second, we extend \approach\ with an active-learning loop, as detailed in Section~\ref{subsec:active_learning}, and then apply it for test-input validation. We  compare  \approach, both with and without active learning, against  state-of-the-art baselines~\cite{Ghobari_2025_active_learning, marsha, WhenWhyTest2022}.

\subsection{Baselines}
\label{subsec:baselines}

We compare \approach\ with two groups of baselines: baselines for test-data labelling and baselines for test-input validation. Both groups of baselines are detailed and described below.

\subsubsection{Test-data labelling baselines} We consider techniques for building highly accurate classifiers, noting that, once trained, these classifiers can automatically label  unlabelled datasets.  In particular, we consider (1) \emph{supervised learning}~\cite{vgg, resnet, vit}, the standard approach for training classifiers, and (2) \emph{pseudo-labelling}~\cite{Pham_2021_MetaPseudoLabel,Lee_2013_PseudoLabel}, a state-of-the-art semi-supervised learning technique that offers an effective strategy for achieving high classification accuracy when most of the training data is unlabelled and only a small portion is labelled. Empirical studies indicate that pseudo-labelling is an effective method across various domains and consistently outperforms established baselines~\cite{Lee_2013_PseudoLabel, Pham_2021_MetaPseudoLabel, Sohn_2021_fixmatch}.
Below, we outline the  approaches underlying  supervised learning and pseudo-labelling. The specific instantiations used in our evaluation are described in Section~\ref{sec:exp-res}.

\textbf{Supervised Learning Baseline.} Algorithm~\ref{alg:supervised} outlines labelling via supervised learning. The algorithm takes as input an unlabelled dataset $D$, the percentage $\mathit{hTotal}$ of $D$ for manual labelling and a classifier $c$, and labels $D$ in three steps. First, $D$ is split into two subsets: a \emph{fine-tuning} subset $D_t$ and a \emph{to-be-labelled} subset $D'$ where $D_t$ comprises a fraction $\mathit{hTotal}$ of $D$ (Line 1 in Algorithm~\ref{alg:supervised}). Then, the fine-tuning subset $D_t$ is manually labelled (Line 2 in Algorithm~\ref{alg:supervised}). Second, the classifier is fine-tuned on $D_t$ (Lines 3 in Algorithm~\ref{alg:supervised}). Third, the fine-tuned classifier labels all the elements  in $D'$ and the labelled set $D_t \cup D'$ is returned.

\small
\begin{algorithm}[t]
\caption{Image Labelling via Supervised Learning~\cite{vgg, resnet, vit}}
\label{alg:supervised}

\footnotesize{
\begin{flushleft}
\textbf{Input} $\mathit{D}$: An unlabelled dataset \\ 
\textbf{Input} $c$: A pretrained classifier\\
\textbf{Param} $\mathit{hTotal}$: Percentage of $D$ to be manually labelled  for supervised learning \\[.5em]
\textbf{Output} $D$: Labelled dataset\\
\end{flushleft}
\begin{algorithmic}[1]
\Statex \CommentG{\textbf{Step 1: Data splitting}} 
\State $(D_t, D') = \text{DataSplit}(D, \mathit{hTotal})$ \CommentG{$D = D_t \cup D', \quad \frac{|D_t|}{|D|} = \mathit{hTotal}$}
\State Manually label $D_t$
\Statex \CommentG{\textbf{Step 2: Fine tuning}} 
\State $\text{FineTune}(c, D_t)$  
\Statex \CommentG{\textbf{Step 3: Labelling $\textbf{D}'$}} 
\For{$\mathit{e}_i \in D'$} 
    \State Let $\hat{l}_i$ be the predicted label for $\mathit{e}_i$ by $c$
    \State Label($\mathit{e}_i$) = $\hat{l}_{i}$  
\EndFor
\State \Return   the labelled set $D$
\end{algorithmic}
}
\end{algorithm}
\normalsize

\textbf{Pseudo-labelling Baseline.} Algorithm~\ref{alg:pseudolabelling} presents the pseudo-labelling approach, as introduced by Lee et al.~\cite{Lee_2013_PseudoLabel} and Pham et al.~\cite{Pham_2021_MetaPseudoLabel}.  The algorithm receives as input an unlabelled dataset $D$, the percentage $\mathit{hTotal}$ of $D$ for manual labelling, the size $v$ of the validation set, and a classifier $c$. It labels $D$ in two steps. First, $D$ is divided into three subsets: a \emph{fine-tuning} subset $D_t$, a \emph{validation} subset $D_v$, and a \emph{to-be-labelled} subset $D'$. The fine-tuning and validation subsets together constitute a proportion $\mathit{hTotal}$ of the input dataset. From this portion, $v$ elements are randomly selected to form the validation set $D_v$, while the remainder is used as the fine-tuning set $D_t$ (Line~1 in Algorithm~\ref{alg:pseudolabelling}). The fine-tuning subset $D_t$ and the validation subset $D_v$ are manually labelled (Line 2 in Algorithm~\ref{alg:pseudolabelling}). Second, the algorithm iteratively generates pseudo-labels -- labels produced by the classifier itself -- for the unlabelled subset \(D'\). It then fine-tunes the classifier on the combined dataset \(D_t \cup D'\), which includes both ground-truth and pseudo-labelled data (Lines~4–9 in Algorithm~\ref{alg:pseudolabelling}). The loop ends either when the classifier begins to overfit or when the time budget is reached. Specifically, Algorithm~\ref{alg:pseudolabelling} uses the validation set $D_v$ to detect overfitting: if $c$'s accuracy on 
 $D_t$ is significantly higher than $c$'s accuracy on $D_v$, it signals overfitting.

\begin{algorithm}[t]
\caption{Image Labelling via Pseudo Labelling~\cite{Pham_2021_MetaPseudoLabel,Lee_2013_PseudoLabel}}
\label{alg:pseudolabelling}

\footnotesize{\begin{flushleft}
\textbf{Input} $\mathit{D}$: An unlabelled dataset \\ 
\textbf{Input} $c$: A pretrained classifier \\
\textbf{Param} $\mathit{hTotal}$: Percentage of $D$ to be manually labelled for semi-supervised learning \\
\textbf{Param} $v$: Number of elements used for validation \\[.5em]
\textbf{Output} $D$: Labelled dataset\\
\end{flushleft}
\begin{algorithmic}[1]

\Statex \CommentG{\textbf{Step 1: Data splitting}} 
\State $(D_t, D_v, D') = \text{DataSplit}(D, \mathit{hTotal}, v)$ \CommentG{$\frac{|D_t \cup D_v|}{|D|} = \mathit{hTotal}, \quad |D_v| = v$}
\State Manually label $D_t$ and $D_v$

\Statex \CommentG{\textbf{Step 2: Iterative pseudo-labelling}} 
\While{$\neg$ OverFitted($c$, $D_t$, $D_v$) \textbf{and} time budget is not over}
    \Statex \CommentG{Generate pseudo labels for $D'$} 
    \For{$\mathit{e}_i \in D'$}
        \State Let $\hat{l}_i$ be the predicted label for $\mathit{e}_i$ by $c$
        \State Label($\mathit{e}_i$) = $\hat{l}_{i}$
    \EndFor
    
    \Statex \CommentG{Fine-tune $c$ using ground-truth–labelled and pseudo-labelled data}
    \State $\text{FineTune}(c, D_t \cup D')$ 
\EndWhile

\State \Return   the labelled set $D$

\end{algorithmic}}
\end{algorithm}
\normalsize

Compared to Algorithm~\ref{alg:supervised}, Algorithm~\ref{alg:pseudolabelling} offers a key advantage by treating pseudo-labels as ground truth for training \(c\). In this way, the absence of ground-truth labels for some data elements does not prevent the classifier from learning from them. Hence, the classifier is trained on a larger, more diverse set of examples.

\subsubsection{Test-input validation baselines}
\label{subsec:baselines-test-input-validation} 
We consider  state-of-the-art techniques from the literature \cite{marsha, WhenWhyTest2022, Ghobari_2025_active_learning}. Each baseline receives a pair of images -- one original and one transformed -- and determines the validity of the transformation using  image-comparison metrics~\cite{stocco}. These metrics quantify how closely a transformed image aligns with its corresponding original. 
Below, we outline these baselines which we  refer to as \emph{B-VIF}~\cite{marsha}, \emph{B-VAE}~\cite{WhenWhyTest2022}, \emph{B-HiL-TV1}~\cite{Ghobari_2025_active_learning}, and \emph{B-HiL-TV2}~\cite{Ghobari_2025_active_learning}:

\textbf{B-VIF Baseline~\cite{marsha}}: The B-VIF baseline computes the Visual Information Fidelity (VIF) metric~\cite{vif} to compare each transformed image against its original.

Given a dataset, B-VIF uses a manually pre-validated subset to select an optimized threshold for the VIF values to distinguish between valid and invalid pairs.
It then uses the threshold to automatically validate the remaining images in the dataset. The manual effort of this baseline is then the size of the manually pre-validated subset, and its accuracy is measured by how well it classifies the remaining data automatically.

\textbf{B-VAE Baseline~\cite{WhenWhyTest2022}}: The B-VAE baseline uses the variational autoencoder (VAE) reconstruction-error approach of Riccio and Tonella \cite{WhenWhyTest2022} as its image-comparison metric. To compute this metric, a VAE is trained on the original images in a given dataset to learn their distribution. Provided with a trained VAE, B-VAE automatically validates each transformed image by computing its reconstruction error and comparing the error against an optimized threshold. As with the B-VIF baseline, this method uses a manually pre-validated subset of images to determine the optimized threshold that separates valid from invalid pairs. The manual effort therefore equals the size of this pre-validated subset, and the baseline's accuracy is evaluated by how well it automatically classifies the remaining data.

\textbf{B-HiL-TV1 Baseline~\cite{Ghobari_2025_active_learning}}: The B-HiL-TV1 baseline uses 13 image-comparison metrics, including VIF and VAE reconstruction error -- used respectively by the B-VIF and B-VAE baselines already described -- to validate transformed images.
Unlike B-VIF and B-VAE, which use a single metric, B-HiL-TV1 trains a random-forest classifier on all the 13 computed metrics to classify transformed images as valid or invalid. Specifically, it uses a manually pre-validated subset of images to train the random-forest classifier. The classifier is then applied to the remaining data.
The manual effort is determined by the size of the pre-validated subset, and the baseline’s accuracy is evaluated based on how well it  classifies the remaining data.

\textbf{B-HiL-TV2 Baseline~\cite{Ghobari_2025_active_learning}}:  The B-HiL-TV2 baseline extends B-HiL-TV1 with an active-learning loop. Like B-HiL-TV1, B-HiL-TV2 uses a manually pre-validated dataset to train a random-forest classifier. Within the active-learning loop, the random-forest classifier labels transformed images as valid or invalid. Based on a confidence threshold for the  predictions, only high-confidence predictions are accepted. A small number of images with low-confidence predictions are selected for manual validation. B-HiL-TV2 then retrains the random-forest classifier by adding the newly labelled images to the existing training set and repeats the active-learning loop until all transformed images are validated. The manual effort for this baseline includes both the pre-validated subset and the images selected for manual validation in each iteration of active learning. B-HiL-TV2 has two user-specified parameters: (1) the confidence threshold used to determine when predictions from the random-forest classifier are accepted, and (2) the maximum number of images sent for manual validation in each iteration of active learning.

\subsection{Datasets}
\label{subsec:datasets}
Table~\ref{table:datasets} summarizes the properties of the 
nine
datasets used in our evaluation. \textsc{CIFAR-10}~\cite{cifar10}, \textsc{FashionMNIST}~\cite{fashionmnist}, \textsc{MNIST}~\cite{mnist}, \textsc{SVHN}~\cite{svhn} 
and \textsc{CelebA}~\cite{celeba}
are widely used benchmarks containing real-world images. \textsc{Synthetic-Pub1} and \textsc{Synthetic-Pub2}~\cite{marsha} are public datasets with synthetically generated images, while \textsc{Industry} is a proprietary industrial dataset also consisting of synthetic images.

\begin{table}[t]
\begin{center}
{\footnotesize
\caption{Properties of the datasets used in our evaluation}
\label{table:datasets}

\scalebox{0.9}{\begin{tabular}{|p{4.5cm}|C{2cm}|C{2cm}|C{2cm}|C{1cm}|}
\hline
\textbf{Dataset} & \textbf{Dataset Size} & \textbf{Public / Proprietary} & \textbf{Real-World / Synthetic} & \textbf{No of Classes} \\
\hline
{\cellcolor{cyan!20}\textsc{Cifar10}}~\cite{cifar10}              & 60,000   & Public       & Real-World   & 10 \\ \hline
{\cellcolor{cyan!20}\textsc{FashionMNIST}}~\cite{fashionmnist}         & 70,000   & Public       & Real-World   & 10 \\ \hline
{\cellcolor{cyan!20}\textsc{MNIST}}~\cite{mnist}                & 70,000   & Public       & Real-World   & 10 \\ \hline
{\cellcolor{cyan!20}\textsc{SVHN}}~\cite{svhn}                 & 73,257  & Public       & Real-World   & 10 \\ \hline
{\cellcolor{cyan!20}\textsc{CelebA-Hair}}~\cite{celeba}                 & 202,599  & Public       & Real-World   & 5 \\ \hline
{\cellcolor{cyan!20}\textsc{CelebA-M/F}}~\cite{celeba}                 & 202,599  & Public       & Real-World   & 2 \\ \hline
{\cellcolor{green!20}\textsc{Synthetic-Pub1}}~\cite{marsha} & 2,318   & Public       & Synthetic    & 2  \\ \hline
{\cellcolor{green!20}\textsc{Synthetic-Pub2}}~\cite{marsha} & 5,204    & Public       & Synthetic    & 2  \\ \hline
{\cellcolor{purple!30}\textsc{Industry}}~\cite{Ghobari_2025_active_learning}           & 3,000    & Proprietary  & Synthetic    & 2  \\ \hline
\end{tabular}}}
\end{center}
\flushleft
\footnotesize{
$^\ast$ The public datasets highlighted with \colorbox{cyan!20}{\makebox(1,1){}} are pre-labelled; the synthetic, public datasets highlighted with \colorbox{green!20}{\makebox(1,1){}} are labelled via crowd-sourcing (Section~\ref{subsec:datasets}); and the synthetic, industry dataset highlighted with \colorbox{purple!30}{\makebox(1,1){}} is labelled by third-party manual labellers and the labels are reviewed by industry experts (Section~\ref{subsec:datasets})}
\vspace*{-.2cm}
\end{table}

Below, we outline the key characteristics of our  datasets, along with how we obtain ground-truth labels for those that were not pre-labelled. In our experiments, we simulate manual labelling by using ground-truth labels. As such, for the purpose of evaluation, ground-truth labels are required for the entire datasets used in our experiments, though outside this context, \approach\ only needs a subset of images to be manually labelled—specifically, the $D_t \cup D_o$ set and the $D_m$ set requested for labelling on Line 17 of Algorithm~\ref{alg:optimzationlabel}.

\textbf{\textsc{CIFAR-10}~\cite{cifar10}} contains $32\times32$ coloured real-world images in 10 different classes, each class comprising 6,000 images, totalling 60,000 images. The classes include common objects such as automobile, airplane, bird, etc.  

\textbf{\textsc{Fashion MNIST}~\cite{fashionmnist}}  contains $28\times28$ grayscale real-world images of fashion items in 10 classes, with each class comprising 7,000 images, totalling 70,000 images. The classes include clothing and accessories such as t-shirt, dress, sneaker, etc.

\textbf{\textsc{MNIST}~\cite{mnist}} contains $28\times28$ grayscale images of hand-written digits in 10 classes, with each class comprising 7,000 images, totalling 70,000 images. The digits range from 0 to 9, commonly used for digit recognition tasks.

\textbf{\textsc{SVHN}~\cite{svhn}}  contains $32\times32$ colour images of house numbers in 10 classes, each class comprising varying numbers of samples, totalling over 73,000 images. The dataset consists of real-world house-number images obtained from Google Street View.

\textbf{\textsc{CelebA}~\cite{celeba}} consists of 202,599 coloured images of celebrity faces, each of size $178 \times 218$, annotated with 40 binary facial attributes.
The images exhibit substantial variation in pose, background, lighting conditions, and facial characteristics, making the dataset a widely used benchmark for facial analysis tasks. In our experiments, we use CelebA for two distinct classification tasks and treat it as two separate datasets. First, we perform a binary classification task to distinguish between \emph{male} and \emph{female} faces, using the corresponding gender attribute provided by the dataset annotations; we denote this task as \textsc{CelebA-M/F}. Second, we formulate a multi-class classification task for hair colour recognition, where images are categorized into five classes: \emph{Black}, \emph{Blond}, \emph{Brown}, \emph{Gray}, and \emph{Bald}, based on the hair-related attribute annotations; we denote this task as \textsc{CelebA-Hair}.

\textbf{\textsc{Synthetic-Pub1} and \textsc{Synthetic-Pub2}} are obtained from the public datasets developed by Hu et al.~\cite{marsha}. Their objective was to create synthetic images for testing vision-based systems. To achieve this, they applied four safety-related image transformations -- \textit{brightness}, \textit{frost}, \textit{contrast}, and \textit{JPEG compression} -- using the Albumentations library~\cite{albumentations} on images from the CIFAR-10~\cite{cifar10} and ImageNet~\cite{imagenet} datasets. Specifically,  from both CIFAR-10 and ImageNet, they selected the  car-class images and an equal number of images from other classes (e.g., cat and ship classes), treating them as the not-car class.  The transformed images may either preserve their essential content, remaining recognizable by humans as having the same label as their corresponding original images, or they may lose their meaning entirely. To determine the labels of the transformed images, Hu et al. presented each transformed image to five human labellers, who were asked to label each transformed image as ``car'' or ``not car''.   We performed a quality review of the images in Hu et al.'s datasets and removed images with multiple inconsistent labels from human labellers when the inconsistencies could not be resolved by majority voting. In the end, we obtained 1,574 unique labelled, synthesized images in \textsc{Synthetic-Pub1} based on the transformed images from  CIFAR-10 and 4,034 unique labelled, synthesized images in \textsc{Synthetic-Pub2} based on the transformed images from ImageNet.

\textbf{\textsc{Industry}} is a proprietary dataset developed in collaboration with an industry partner. This dataset contains synthetic images generated for robustness testing of vision-based anomaly detection solutions for identifying faults in electric transmission lines. The dataset includes images synthesized from real-world pictures of utility poles using a prompt-based generative transformer to depict different climatic conditions that were not represented in the original, real-world images. To label the transformed images, two independent labellers, who are not co-authors of this article, were trained using detailed guidelines provided by our partner. 
They were then asked to label the images as either ``valid'' or ``invalid''. We recall from the example in Section~\ref{subsec:testing_vision_systems} that in this dataset, following the validity definition of Riccio and Tonella~\cite{WhenWhyTest2022}, a ``valid'' image is one that is within the real-world image domain, i.e., confidently recognized as a utility pole by a domain expert, whereas an ``invalid'' image either does not represent a utility pole or is too distorted to be confidently identified as one.
After removing the images where the labellers disagreed on the classification,  3,000 labelled images were reviewed and verified by experts from our industry partner and included in our \textsc{Industry} dataset.

We use all nine datasets to evaluate \approach\ for 
test-data labelling. For test-input validation,  three datasets -- \textsc{Synthetic-Pub1}, \textsc{Synthetic-Pub2}, and \textsc{Industry} -- are applicable, as they contain image pairs. The remaining six datasets do not include image pairs and are therefore not applicable to the test-data validation use case.

\subsection{Evaluation Metrics}\label{subsec:metrics}
\label{subsec:metrics}
We evaluate \approach\ and our baselines using three metrics. The first metric is \textbf{accuracy}, defined  as the percentage of correctly labelled images in an entire dataset. Specifically, for Algorithms~\ref{alg:optimzationlabel},~\ref{alg:supervised}, and~\ref{alg:pseudolabelling}, accuracy is calculated as the percentage of images correctly labelled --either automatically or manually -- in dataset $D$.

The second metric is  \textbf{manual labelling effort}, defined as the ratio of the number of manually labelled images to the size of the entire dataset. As discussed in  Section~\ref{subsec:datasets}, since ground-truth labels are available for all of our experimental datasets, we simulate human labelling by using the corresponding labels for each image from the ground truth. This approach, commonly used in the evaluation of image classifiers ~\cite{Ghobari_2025_active_learning, Chetan_2019_ActiveLearning}, eliminates the need for direct human input. For \approach, the manual labelling effort is the sum of the percentage of initially labelled images, $\mathit{hInitial}$, and the size of set $D_m$, which  is manually labelled on Line 17 of Algorithm~\ref{alg:optimzationlabel}, divided by the size of $D$. For the baselines presented in Algorithms~\ref{alg:supervised} and~\ref{alg:pseudolabelling}, the manual labelling effort is simply the percentage of labelled images, $\mathit{hTotal}$, which is an input to these algorithms.

The last metric is the \textbf{execution time} of each labelling technique. Although neither \approach\ nor the baselines rely on user feedback during their main labelling loop -- since manual labelling occurs either before or after automated labelling --  execution time still has practical implications. Execution time is particularly important when engineers need to label large datasets or frequently update labels in response to data changes. Therefore, we consider execution time to assess and compare the efficiency of \approach\ and the baselines.

\subsection{Implementation}
\label{subsec:imp}
The experiments were conducted on a machine with two Intel Xeon Gold 6338 CPUs, 512 GB of RAM, and one NVIDIA A40 GPU (46~GB memory).
We have implemented \approach\ in Python 3.11 using Pytorch~\cite{pytorch} for classifiers, PuLP~\cite{Mitchell_2011_pulp} for MILP solver and Scikit-learn~\cite{sklearn} for data splitting and clustering.

\section{Experiments and Results}
\label{sec:exp-res}
In this section, we present the experiments performed based on the setup described in Section~\ref{sec:eval} to answer RQ1 and RQ2.

\textbf{RQ1 experiments.} To address RQ1, we run Algorithm~\ref{alg:optimzationlabel} with $\alpha = 100\%$ for each of the 
nine
datasets in Table~\ref{table:datasets}. Configuring \approach\ requires selecting which classifiers (and how many) to include, as well as setting the parameters $\mathit{hInitial}$ and $s$. 

For the choice of classifiers, we use the three state-of-the-art vision-based models listed in Table~\ref{table:dl_models}: VGG (a CNN), ResNet (a CNN with residual connections), and ViT (a transformer-based model). These classifiers represent the main deep-learning architectures for vision-based systems. We run \approach\ using different combinations of these classifiers, specifically with one, two, and all three models from Table~\ref{table:dl_models}. We denote these configurations as \approach($n=1$), \approach($n=2$), and \approach($n=3$), respectively. In \approach($n=1$), each of the three models is used individually, producing three separate instances. In \approach($n=2$), we form three pairs (VGG+ResNet, VGG+ViT, and ResNet+ViT). Finally, \hbox{\approach($n=3$)} uses  all the three models together (VGG+ResNet+ViT).
We note that in the experiment design for this research question, we intentionally focused on architectural diversity (VGG/ResNet/ViT) to demonstrate OPAL’s effectiveness when it uses heterogeneous classifiers. 
While further experiments comparing multiple instances of the same architecture versus instances of different architectures could be conducted, such experiments are more informative for analyzing the underlying DL classifiers than for evaluating the performance of the \approach\ optimization framework itself. Such experiments can be considered as a direction for future work.

We vary the $\mathit{hInitial}$ parameter, i.e., the portion of initially labelled data, setting it to 1\%, 5\%, 15\%, 25\%, 35\%, and 45\% of the entire input dataset $D$.  We set  $s$, i.e.,  the size of  \(D_o\), to either 1000 or half of $\mathit{hInitial}$ multiplied by the size of $D$, whichever is smaller. This ensures that the number of constraints in \approach's MILP formulation remains around 1000.  With this value of \(|D_o|\), the MILP's objective function -- i.e., human effort -- starts to saturate, and increasing \(|D_o|\) further does not yield additional improvements in minimizing human effort for the set $D_o$. To mitigate randomness, we repeat each experiment five times.  
Overall, for RQ1, we perform $9 (\textit{Datasets}) \times 7(\textit{DL-model Combinations}) \times 6 (\textit{hInitial}) \times 5 (\textit{Repetitions}) = 1890$ experiments.

\begin{table}[t]
\begin{center}
\caption{Classifiers used as inputs to Algorithms~\ref{alg:optimzationlabel},~\ref{alg:supervised}, and~\ref{alg:pseudolabelling}}
\label{table:dl_models}
\scalebox{0.85}{
\begin{tabular}{|p{2.5cm}|p{3.8cm}|p{2.8cm}|p{3cm}|} 
\hline
\textbf{Classifier} & \textbf{Architecture Type} & \textbf{Depth / Blocks} & \textbf{Model Scale} \\ \hline
VGG~\cite{vgg} 
& CNN (plain convolutional) 
& 16 layers 
& Medium \\ \hline
ResNet~\cite{resnet} 
& CNN (residual connections) 
& 50 layers 
& Large \\ \hline
ViT~\cite{vit} 
& Transformer-based 
& 12 transformer blocks 
& Large \\ \hline
\end{tabular}}
\end{center}
\end{table}

For the data splitting step of \approach, we first represent images in $D$ as latent-space vectors obtained from the pretrained VGG model~\cite{vgg}. 
These vectors are then clustered using the DBSCAN algorithm~\cite{dbscan}. The algorithm's parameters, namely, epsilon and the minimum number of samples per cluster, are configured in accordance with established best practices
and Euclidean norm is used for the distance calculation within the algorithm
~\cite{dbscan, Han_2011_data_mining}.
We note that clustering methods other than DBSCAN, such as K-Means, could be used for the data-splitting step of \approach, since the goal at this stage is simply to sample a diverse and representative subset of images from $D$. This goal can be achieved with any clustering method given appropriate parameter settings. However, methods such as K-Means require predefining the number of clusters, which introduces additional complexity: too few clusters may reduce diversity, while too many clusters may lead to sparsely populated clusters, from which sampling becomes sensitive to minor data variations and may over-represent outliers or noise. Moreover, the appropriate number of clusters is data-dependent and often difficult to determine in advance. For these reasons, we adopt DBSCAN, configured with the best-practice guidelines from the literature~\cite{dbscan}, which automatically determines the number of clusters and improves the diversity and representativeness of the sampled images compared to a completely random sampling method.
 The fine-tuning step of \approach\ uses 20 epochs with batch sizes and learning rates set according to the recommended configurations for each  classifier. The code and data are available~\cite{github}.

\begin{figure}[t]
    \centering
    \includegraphics[width=0.98\linewidth]{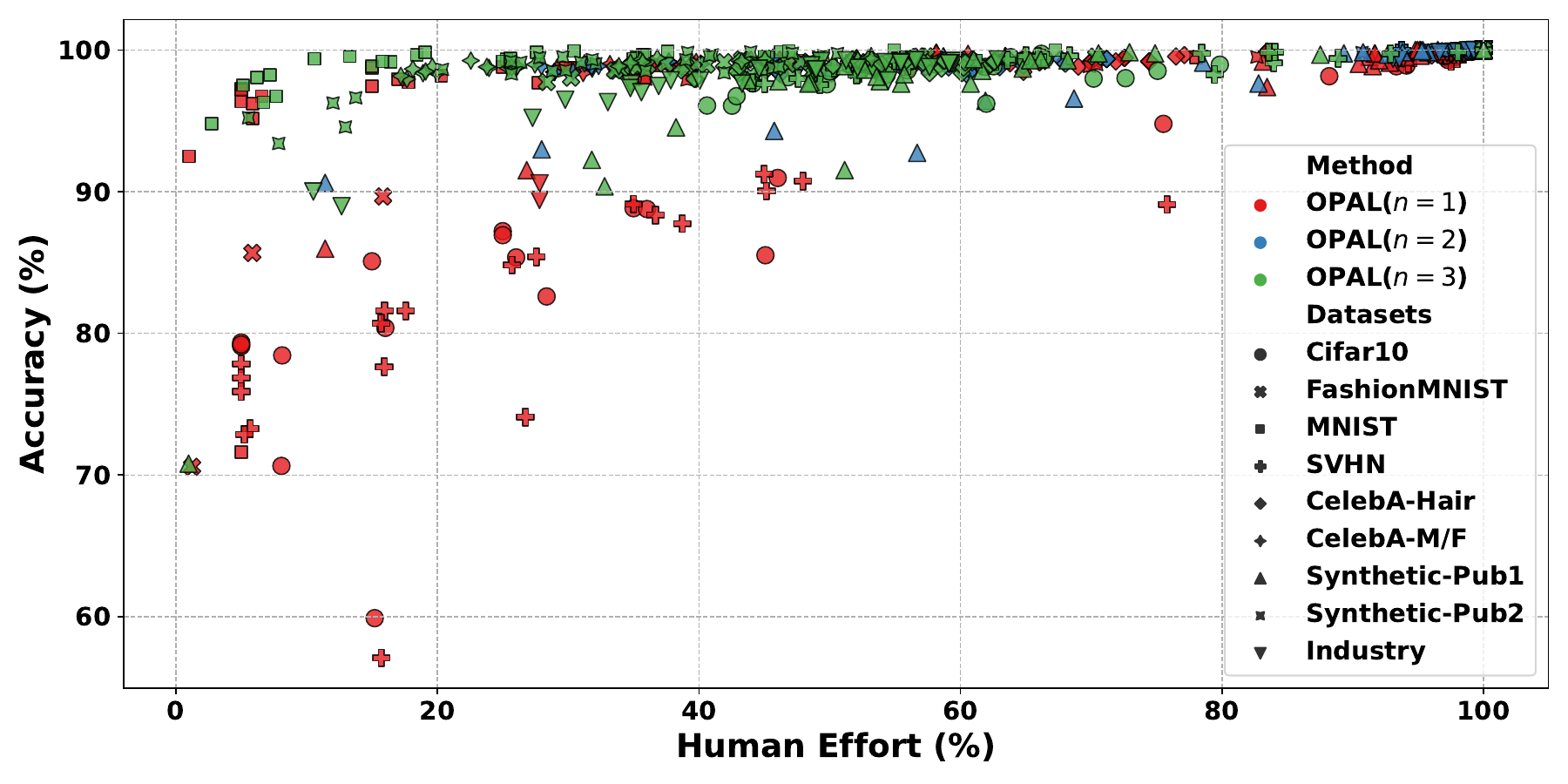}
    \caption{Labelling accuracy vs. manual labelling effort for different configurations of \approach, different portions of manually pre-labelled data (\emph{hInitial}) and different datasets}
    \label{fig:rq1_acc_vs_humaneffort}
\end{figure}

\textbf{RQ1 results.}
Figure~\ref{fig:rq1_acc_vs_humaneffort} shows a scatter plot of accuracy versus manual labelling effort, where each point corresponds to an experiment performed for a specific dataset, a particular value of $\mathit{hInitial}$, and one of the three configurations \approach(\(n=1\)), \approach(\(n=2\)), or \approach(\(n=3\)). We use colours to differentiate between different configurations and shapes to differentiate between different datasets.
Table~\ref{table:rq1_summary} shows average accuracy and manual labelling effort associated with these configurations.
Table~\ref{table:rq1_stat} reports the Vargha-Delaney $\hat{A}_{12}$ effect-size test~\cite{vargha} comparing the accuracy and manual effort values achieved by \approach($n=3$) versus those of \approach($n=1$) and \approach($n=2$). The Wilcoxon Signed-Rank test~\cite{wilcoxon} confirms that \approach($n=3$) significantly outperforms \approach($n=1$) and \approach($n=2$) with p-values $\ll 0.01$.
With a single classifier ($n=1$), \approach\ achieves an average accuracy of  
96.4
while requiring an average manual effort of 
69.3
. Increasing to two classifiers ($n=2$) boosts the average accuracy to 
98.2
and reduces the average manual effort to 
58.7\%
. Finally, using three classifiers ($n=3$) yields the highest average accuracy of 
98.5\%
, while further lowering the average manual effort to  
47.1\%
. The results indicate that increasing the number of classifiers  improves accuracy and substantially reduces the labelling effort. To further investigate the impact of increasing the number of classifiers on \approach's performance, Table~\ref{table:rq1_conditions} shows the average percentage of images that satisfy \textbf{Condition 1} and  \textbf{Condition 2}, both individually and in conjunction, for different number of classifiers. Note that for \approach\ ($n=1$), \textbf{Condition 1} holds vacuously for all images. As the table shows, 
\textbf{Condition~2} is more restrictive than \textbf{Condition~1} and is satisfied by fewer images than \textbf{Condition~1} in all cases.  Given that the rates at which images satisfy \textbf{Condition~2} and the conjunction of these two conditions are close, \textbf{Condition~2} largely determines whether, or not,  an image is labelled automatically.  Increasing the number of classifiers makes \textbf{Condition 2} weaker and easier to satisfy, leading \approach\ to automatically label more images and thereby reducing the amount of manual labelling.

While \approach($n=3$) surpasses \approach(\(n=1\)) and \approach(\(n=2\)) in both accuracy and manual effort, it only modestly increases execution time. Specifically, the average execution times for \(n=1\), \(n=2\), and \(n=3\) are 
544, 
767 and 
926
seconds, respectively.
To calculate these execution times, we measure the time for data splitting, fine-tuning, MILP solving and labelling.
Given \approach($n=3$)'s better performance, we focus on this configuration in the rest of  RQ1 and in RQ2, and in the remainder of Section~\ref{sec:exp-res},  \approach\ refers to \approach($n=3$).


\begin{table}[t]
\centering
\caption{Average percentage of images satisfying Condition~1 and  Condition~2, both individually and together, in Algorithm~\ref{alg:optimzationlabel} as the number of classifiers ($\textbf{n}$) increases}
\label{table:rq1_conditions}
\scalebox{0.95}{
\begin{tabular}{|l|c|c|c|}
\hline
   \textbf{Conditions being fulfilled}     & \textbf{n=1} & \textbf{n=2} & \textbf{n=3} \\ \hline
Condition 1 & All     & 
65.3\%
& 
62.5\%
\\ \hline
Condition 2 &
34.3\%
& 
44.3\%
& 
58.7\%
\\ \hline
Condition 1 and 
Condition 2 &      
34.3\%
&   
41.7\%
&
51.8\%
\\ \hline
\end{tabular}
}
\end{table}

\begin{table}[t]
\centering
\caption{Average manual effort and accuracy of OPAL with different number of classifiers ($n$) across all datasets.}
\label{table:rq1_summary}
\scriptsize
\scalebox{0.95}{
\subfloat[Manual Effort]{
\centering
\setlength{\tabcolsep}{2pt}
\begin{tabularx}{\textwidth}{l*{10}{Y}}
\toprule
\textbf{Method} &
\rot{\textbf{CIFAR-10}} &
\rot{\textbf{Fashion}\\\textbf{MNIST}} &
\rot{\textbf{MNIST}} &
\rot{\textbf{SVHN}} &
\rot{\textbf{CelebA-Hair}} &
\rot{\textbf{CelebA-M/F}} &
\rot{\textbf{Synthetic}\\\textbf{Pub1}} &
\rot{\textbf{Synthetic}\\\textbf{Pub2}} &
\rot{\textbf{Industry}} &
\rot{\textbf{Avg.}} \\
\midrule
\textbf{OPAL($n$=1)} & 0.561 & 0.701 & 0.591 & 0.497 & 0.691 & 0.456 & 0.903 & 0.897 & 0.944 & 0.693 \\
\textbf{OPAL($n$=2)} & 0.699 & 0.546 & 0.373 & 0.535 & 0.604 & 0.402 & 0.692 & 0.562 & 0.872 & 0.587 \\
\textbf{OPAL($n$=3)} & 0.608 & 0.470 & 0.269 & 0.653 & 0.527 & 0.330 & 0.606 & 0.327 & 0.451 & 0.471 \\
\bottomrule
\end{tabularx}
}
}
\vspace{6pt}

\scalebox{0.95}{
\subfloat[Accuracy]{
\centering
\setlength{\tabcolsep}{2pt}
\begin{tabularx}{\textwidth}{l*{10}{Y}}
\toprule
\textbf{Method} &
\rot{\textbf{CIFAR-10}} &
\rot{\textbf{Fashion}\\\textbf{MNIST}} &
\rot{\textbf{MNIST}} &
\rot{\textbf{SVHN}} &
\rot{\textbf{CelebA-Hair}} &
\rot{\textbf{CelebA-M/F}} &
\rot{\textbf{Synthetic}\\\textbf{Pub1}} &
\rot{\textbf{Synthetic}\\\textbf{Pub2}} &
\rot{\textbf{Industry}} &
\rot{\textbf{Avg.}} \\
\midrule
\textbf{OPAL($n$=1)} & 0.899 & 0.972 & 0.976 & 0.874 & 0.992 & 0.991 & 0.979 & 0.997 & 0.993 & 0.964 \\
\textbf{OPAL($n$=2)} & 0.981 & 0.976 & 0.982 & 0.936 & 0.989 & 0.990 & 1.000 & 0.986 & 0.999 & 0.982 \\
\textbf{OPAL($n$=3)} & 0.985 & 0.989 & 0.990 & 0.990 & 0.989 & 0.990 & 0.970 & 0.985 & 0.977 & 0.985 \\
\bottomrule
\end{tabularx}
}
}
\end{table}

\begin{table}[t]
\caption{Vargha-Delaney $\hat{\textbf{A}}_{\textbf{12}}$ statistical test between \approach(n=3) versus \approach(n=2) and \approach(n=1) for both manual effort and accuracy. In all comparisons, we have p-value $\ll$ 0.01}
\label{table:rq1_stat}
\begin{center}
\scalebox{0.95}{
\begin{tabular}{|c|c|c|c|}
\hline
\multicolumn{2}{|c|}{vs OPAL (n=1)} & \multicolumn{2}{c|}{vs OPAL (n=2)} \\
\hline
Effort & Accuracy & Effort & Accuracy \\
\hline
0.67 (M) & 0.6 (S) & 0.89 (L) & 0.85 (L) \\
\hline
\end{tabular}
}
\end{center}
\end{table}

\begin{figure}[t]
    \centering
    \includegraphics[width=0.98\linewidth]{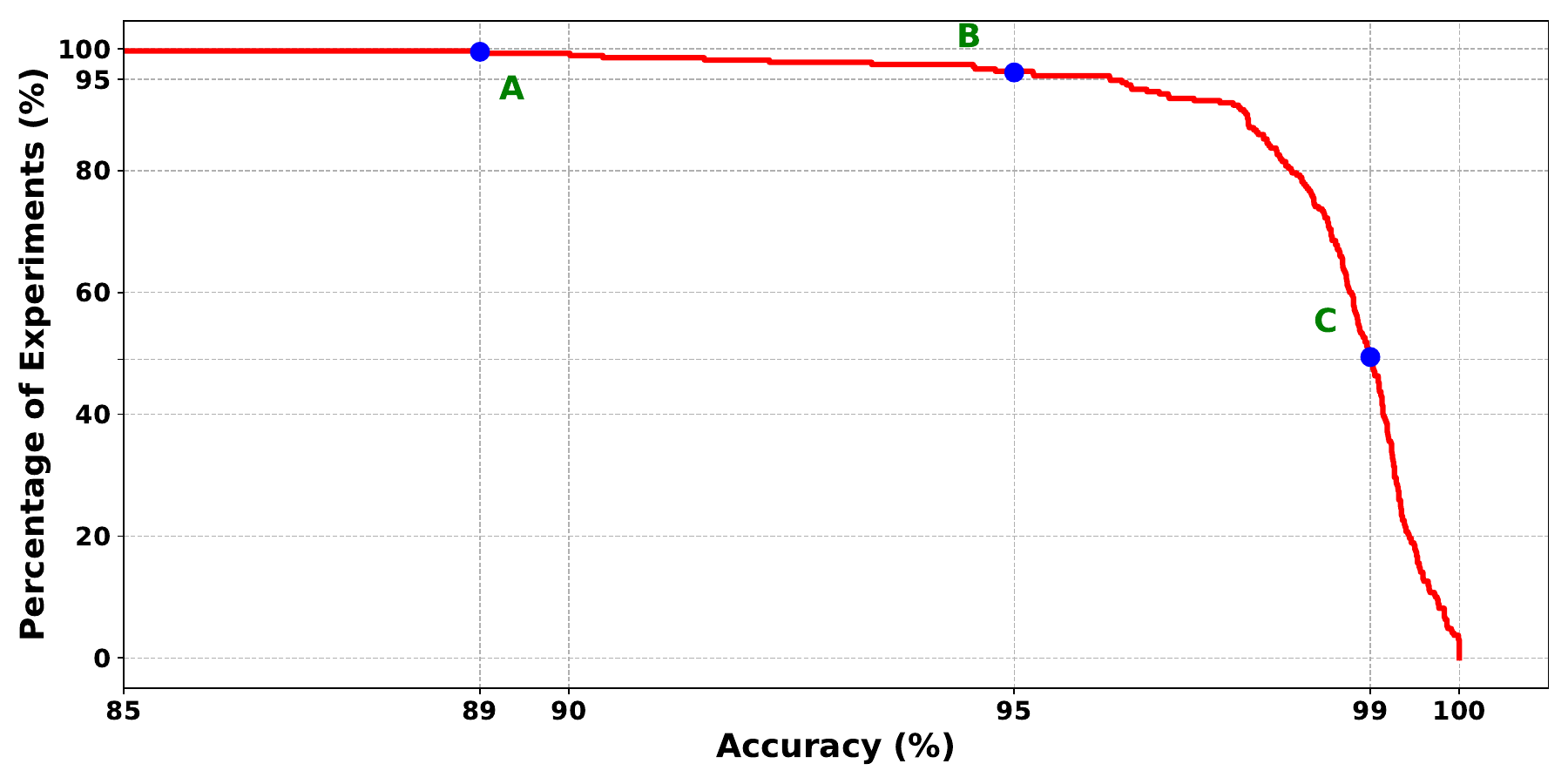}

    \caption{The proportion of \approach's experiments with \textit{n=3}  from Figure~\ref{fig:rq1_acc_vs_humaneffort} that achieve an accuracy at or above a specified threshold: Point \textbf{(A)} shows 99.5\% of the experiments
    achieve at least  
    89\%
    accuracy; point \textbf{(B)} shows 
    96.1\%
    of the experiments achieve at least 95\% accuracy; and point  \textbf{(C)} shows  
    51.1\%
    of the experiments achieve at least   99\% accuracy}
    \label{fig:rq1_acc_survival}
\end{figure}

To examine the probability that \approach($n=3$) achieves a labelling accuracy above a given threshold, we present the complementary cumulative distribution function (i.e., \(1 - \mathrm{CDF}\)) in Figure~\ref{fig:rq1_acc_survival} computed based on the accuracy results for \approach($n=3$) from  Figure~\ref{fig:rq1_acc_vs_humaneffort}. The three points highlighted in  Figure~\ref{fig:rq1_acc_survival} summarize the accuracy performance of  \approach($n=3$)  as follows: \textbf{(A)}  In 99.5\% of the experiments, \approach{} achieves at least 
89\%
accuracy. \textbf{(B)}  In 
96.1\%
of the experiments, \approach{} achieves at least 95\% accuracy.  \textbf{(C)} In 
51.1\%
of the  experiments, \approach\ achieves a minimum accuracy of 99\%.

\begin{figure}[t]
    \centering
    \includegraphics[width=0.7\linewidth]{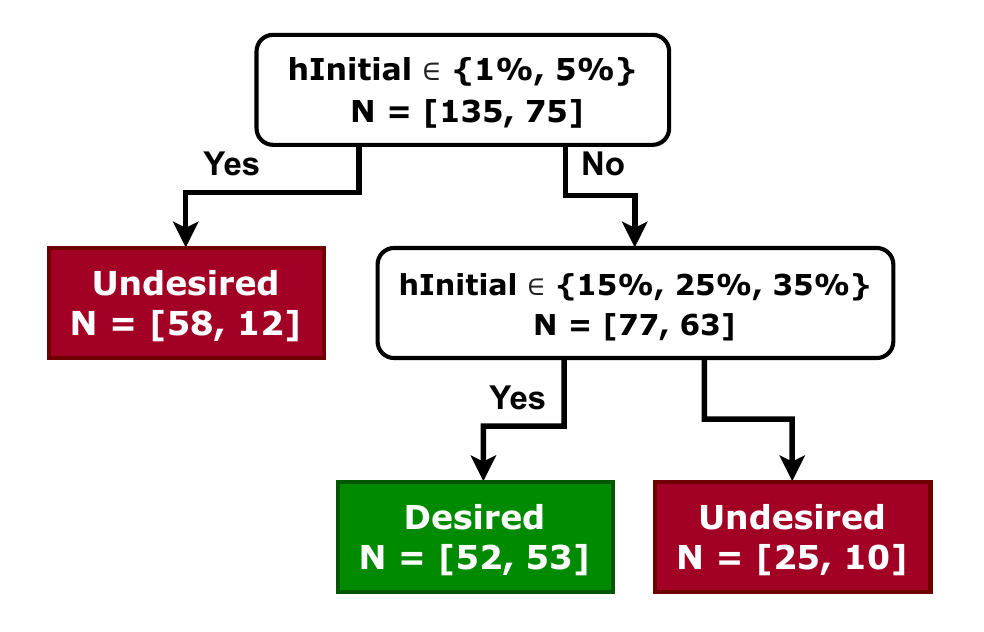}
    \caption{Decision tree for manual pre-labelling budget (\emph{hInitial}) in \approach: The decision tree identifies \emph{hInitial} values  that yield our desired performance, achieving at least 98\% accuracy while keeping manual effort below 50\%. At each node, N = [x, y] indicates the count of undesired (x) and desired (y) experiments reaching that point}
    \label{fig:dt}
\end{figure}

We next explore how to configure the parameter \(\mathit{hInitial}\). We use decision-tree analysis to identify $\mathit{hInitial}$ values that achieve at least 98\% accuracy while keeping manual effort below 50\%. 
These thresholds represent practical trade-offs in real-world applications
as suggested by recent industrial research~\cite{Ghobari_2025_active_learning}.
To build a decision tree that generalizes effectively across multiple datasets, we aggregate the \approach(\(n=3\)) results from all nine datasets in Table~\ref{table:datasets}. We then split the combined dataset into two groups: data points with accuracy \(\geq\) 98\% and manual effort \(\leq\) 50\% are considered ``desired'', while all other data points are deemed ``undesired''. Figure~\ref{fig:dt} shows the resulting  decision tree. At each node, the tree identifies the most influential value of $\mathit{hInitial}$ for classifying data points as desired or undesired. As the tree shows, the $\mathit{hInitial}$ values of 15\%, 25\% and 35\% are more likely to yield the desired trade-offs than the values lower than $15$\% and higher than $35$\%. For these $\mathit{hInitial}$ values, \approach($n=3$) achieves an average accuracy of 98.8\%  while requiring only 
44.2\%
manual effort.

\begin{figure}[t]
    \centering
    \includegraphics[width=0.98\linewidth]{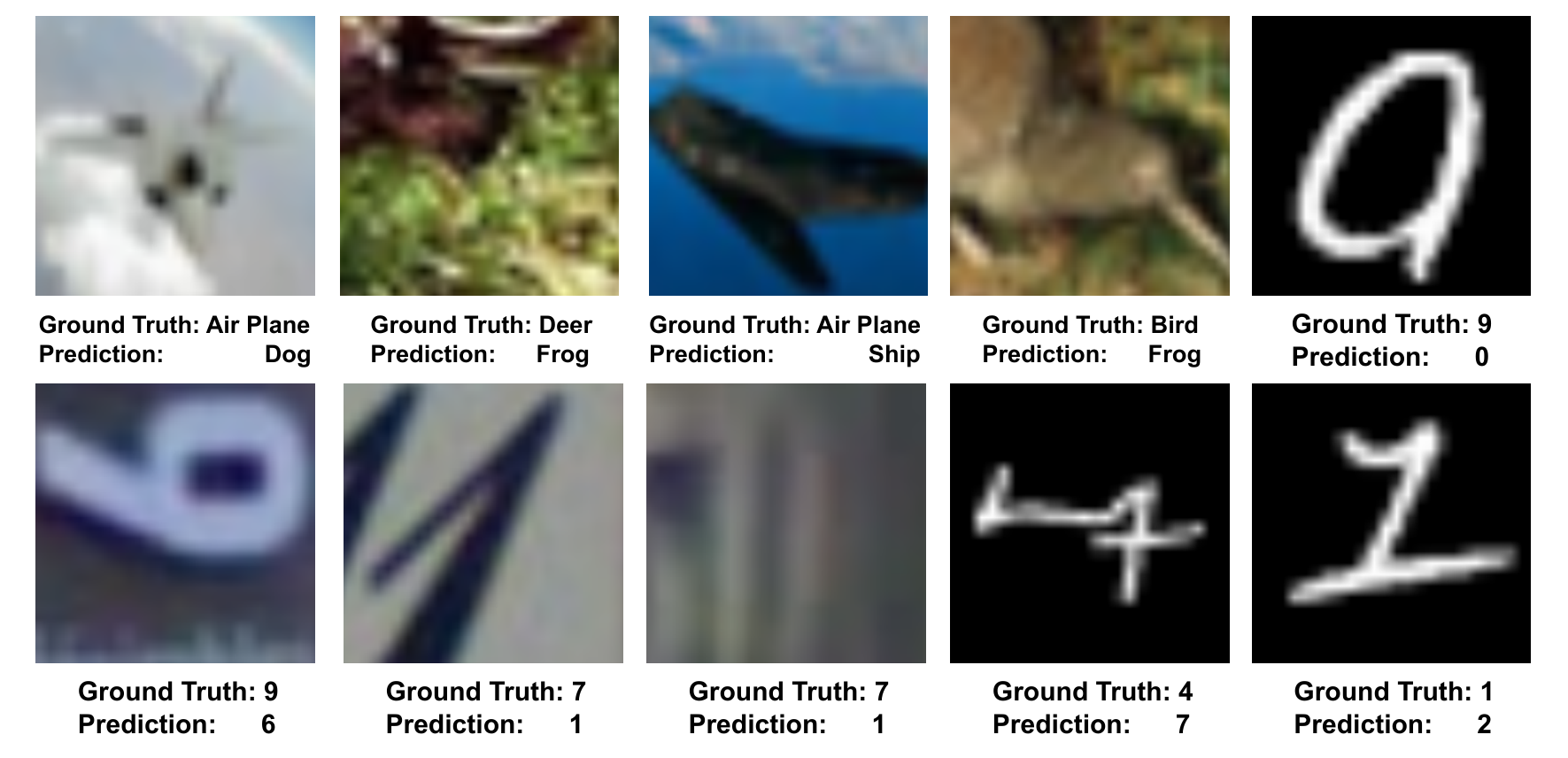}
    \caption{Examples of images from MNIST, SVHN, \textsc{Synthetic-Pub1}, and \textsc{Synthetic-Pub2} that were mislabelled by both \approach\ and the human labeller, along with their ground-truth labels and the predictions made by \approach}
    \label{fig:misclassifications}
    \vspace*{-.4cm}
\end{figure}

We engaged an independent labeller -- who is not an author of this study -- to qualitatively analyze the images mislabelled by \approach.
The independent labeller was a graduate student with background in computer vision who was not involved in the design or implementation of \approach. The labeller was provided with an introduction to the datasets used in the study and was asked to assign labels to the selected samples in an offline setting, and was compensated at a fixed hourly rate. The purpose of this experiment was to provide an external \textit{qualitative assessment} of \approach’s final labelling decisions rather than to conduct a full inter-annotator agreement study.
We collected all mislabelled images from a single run in which \approach\ was applied using three classifiers with \(\mathit{hInitial} = 15\%\) across our nine datasets.  This resulted in a set of 1,936 images -- constituting  0.6\% (six-tenths of a percent) of all images in our datasets -- being mislabelled by \approach.  We provided the labeller with the descriptions of our public datasets (see Section~\ref{subsec:datasets}) so that they knew the class labels for each dataset and requested that they label the images mislabelled by \approach.   The labeller was not aware of \approach’s labels for these images.  Out of the 1,936 images, the human labeller also misclassified 561 images -- about 29\% -- according to the ground truth. These images are available in our replication package~\cite{github}. Figure~\ref{fig:misclassifications} shows some examples of images from MNIST, SVHN, \textsc{Synthetic-Pub1}, and \textsc{Synthetic-Pub2} that were mislabelled by both \approach\ and the human labeller, along with their ground-truth labels and the predictions made by \approach. Although these problematic images make up well under 1\% of the total, this finding suggests that perfect labelling accuracy may be unattainable even for human annotators due to low data quality in some datasets and the inherent subjectivity of labelling complex or low-quality images.
However, we note that due to \approach's design, images that do not meet \approach's agreement-and-confidence conditions are explicitly routed to human annotators, where the annotators may provide a label or flag the sample as invalid/out-of-distribution and discard it.

\begin{tcolorbox}[breakable, colback=gray!10!white,colframe=black!75!black]

\textbf{Finding:} Based on experiments across 
nine
diverse datasets, when \approach\ is configured to target 100\% accuracy, it achieves an average accuracy of 98.8\%, just 1.2\% below perfect accuracy, while requiring 
44.2\%
manual effort, effectively cutting manual labelling by more than half.

\textbf{Takeaway~1:} When applying \approach\ to vision-based classification tasks, we recommend combining multiple vision-based classifiers with different architectures -- such as CNNs and transformers. Our results -- consistently observed across 
nine
datasets --  show that using three distinct CNN-based and transformer-based models significantly improves accuracy and reduces manual effort compared to relying on just one or two.

\textbf{Takeaway~2:}  Using only 1–5\% of pre-labelled data for fine-tuning and optimization in \approach\ yields low accuracy. Conversely, using 45\% of pre-labelled data pushes manual effort above 50\%, while accuracy improves only marginally. The best trade-off between accuracy and manual effort is achieved when 15–35\% of the data are pre-labelled (i.e., when \emph{hInitial} is set to 15–35\%).

\textbf{Takeaway~3:} Among images misclassified by \approach, approximately $30$\% were also  misclassified by a human annotator, indicating that perfect labelling accuracy may be unattainable even for human labellers. 

\end{tcolorbox}

\textbf{RQ2 experiments.}
To compare \approach\ with baselines, i.e., Algorithms~\ref{alg:supervised} and~\ref{alg:pseudolabelling}, we apply the baselines to the datasets in Table~\ref{table:datasets} and compare their results with \approach's results from RQ1. Configuring Algorithms~\ref{alg:supervised} and~\ref{alg:pseudolabelling} requires selecting a classifier and setting the parameter $\mathit{hTotal}$. Algorithm~\ref{alg:pseudolabelling} further requires setting parameter $v$, i.e., the size of the validation set. 

We apply the supervised learning method,  i.e., Algorithm~\ref{alg:supervised}, to each classifier in Table~\ref{table:dl_models}, referring to the resulting baselines as \textit{B-VGG}, \textit{B-ResNet}, and \textit{B-ViT}. For the pseudo-labelling method, we execute Algorithm~\ref{alg:pseudolabelling} exclusively with the VGG model since, in our preliminary experiments, VGG demonstrates the best performance among the three. Note that the pseudo-labelling baseline is substantially more expensive than supervised learning, as Algorithm~\ref{alg:pseudolabelling} fine-tunes the classifier multiple times in a loop over the entire dataset. Hence, we only evaluate the pseudo-labelling baseline with VGG, referring to it as \textit{B-Pseudo}.

\begin{figure*}[t]
    \centering
    \includegraphics[width=\linewidth]{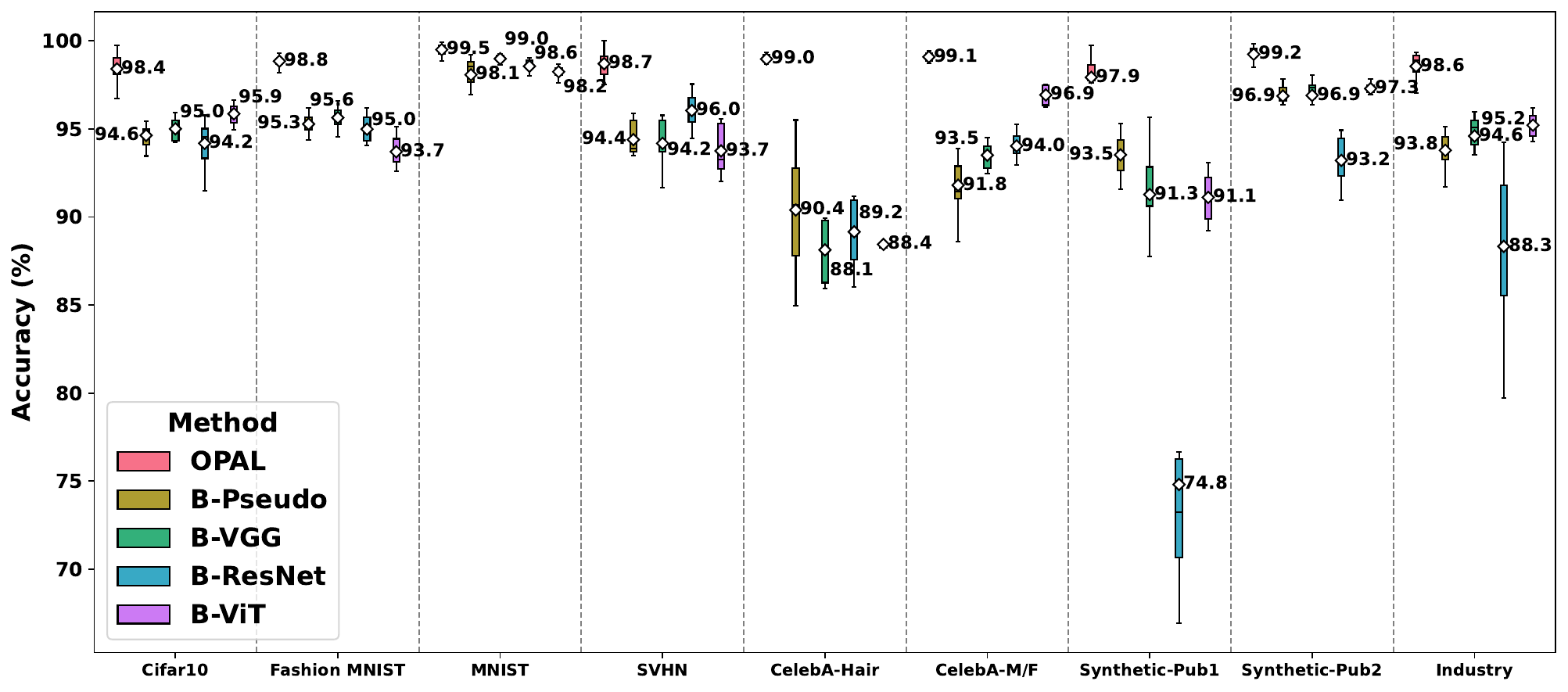}

    \caption{Accuracy distributions obtained by \approach\ and the baselines when all methods are provided with the same manual labelling effort that \approach\ requires in RQ1 to achieve near-perfect accuracy}
    \label{fig:rq2_boxplot}
\end{figure*}

\begin{table}[t]
\centering
\caption{Average accuracy obtained by \approach\ and the baselines with the same manual labelling effort as \approach\ requires in RQ1 for each dataset to achieve near-perfect accuracy.}
\label{table:rq2_summary}
\scriptsize
\setlength{\tabcolsep}{3pt}

\begin{tabularx}{\textwidth}{l*{9}{Y}}
\toprule
\textbf{Method} &
\rot{\textbf{CIFAR-10}} &
\rot{\textbf{Fashion}\\\textbf{MNIST}} &
\rot{\textbf{MNIST}} &
\rot{\textbf{SVHN}} &
\rot{\textbf{CelebA}\\\textbf{Hair}} &
\rot{\textbf{CelebA}\\\textbf{M/F}} &
\rot{\textbf{Synthetic}\\\textbf{Pub1}} &
\rot{\textbf{Synthetic}\\\textbf{Pub2}} &
\rot{\textbf{Industry}} \\
\midrule
\textbf{OPAL}     & 98.4 & 98.8 & 99.5 & 98.7 & 99.0 & 99.1 & 97.9 & 99.2 & 98.6 \\
\textbf{B-Pseudo} & 94.6 & 95.3 & 98.1 & 94.4 & 90.4 & 91.8 & 93.5 & 96.9 & 93.8 \\
\textbf{B-VGG}    & 95.0 & 95.6 & 99.0 & 94.2 & 88.1 & 93.5 & 91.3 & 96.9 & 94.6 \\
\textbf{B-ResNet} & 94.2 & 95.0 & 98.6 & 96.0 & 89.2 & 94.0 & 74.8 & 93.2 & 88.3 \\
\textbf{B-ViT}    & 95.9 & 93.7 & 98.2 & 93.7 & 88.4 & 96.9 & 91.1 & 97.3 & 95.2 \\
\bottomrule
\end{tabularx}
\end{table}

The parameter $\mathit{hTotal}$ in Algorithms~\ref{alg:supervised} and~\ref{alg:pseudolabelling} specifies the total amount of manual labelling effort required for  supervised learning and pseudo-labelling, respectively.  To ensure a fair comparison between \approach\ and the baselines, i.e., that all methods operate under the same total manual labelling effort, we first calculate the average manual effort \approach\ requires for each dataset in Table~\ref{table:datasets}, using the results from RQ1. This process yields six average manual effort values for each dataset, each corresponding to one of the six different levels of pre-labelled data -- that is, the six different values for $\mathit{hInitial}$, considered in RQ1 experiments.  For each dataset, we  set the parameter $\mathit{hTotal}$ in Algorithms~\ref{alg:supervised} and~\ref{alg:pseudolabelling} to these six values, respectively. This ensures that \approach\ and the baselines are evaluated under an identical total manual labelling budget. We set $v$ in Algorithm~\ref{alg:pseudolabelling} to 15\% of the manually labelled elements, allocating the  remaining 85\%  for fine-tuning -- a common validation/fine-tuning split in the DL community~\cite{Goodfellow_2016_DLBook}. To maintain consistency, we implement the data splitting and fine-tuning steps in both Algorithms~\ref{alg:supervised} and~\ref{alg:pseudolabelling} identically to \approach. To address randomness, as in RQ1, we repeat each experiment five times. In total, for RQ2, we perform 
$4 (\textit{baselines}) \times 6 (\textit{labelling budgets}) \times 9 (\textit{datasets}) = 216$
experiments.

\textbf{RQ2 results.} Figure~\ref{fig:rq2_boxplot} shows the labelling accuracy of \approach\ compared to the four baselines -- \textit{B-VGG}, \textit{B-ResNet}, \textit{B-ViT}, and \textit{B-Pseudo} -- under the same manual labelling budget. 
The average accuracy for each method-dataset pair is reported in Table~\ref{table:rq2_summary}.
The Wilcoxon Signed-Rank test~\cite{wilcoxon} confirms that \approach's accuracy significantly outperforms that of all baselines (p-value $\ll 0.01$) across all datasets.  Table~\ref{table:rq2_stat} presents the results of the Vargha-Delaney $\hat{A}{12}$ effect size test~\cite{vargha}, comparing the accuracy of \approach\ against each baseline across all datasets. According to the Vargha and Delaney classification of effect sizes, \(\hat{A}_{12}\) values of 0.56, 0.64, and 0.71 or higher correspond to small (S), medium (M), and large (L) effect sizes, respectively~\cite{vargha}. Given  these thresholds, \approach\ consistently demonstrates a large effect size on every dataset, indicating that it significantly outperforms all baselines for all datasets.

\begin{table}[t]
\centering
\caption{Vargha-Delaney $\hat{\textbf{A}}_{\textbf{12}}$ statistical test between \approach's labelling accuracy versus baselines. In all comparisons, we have p-value $\ll$ 0.01}
\label{table:rq2_stat}
\scalebox{0.85}{
\begin{tabular}{|l|c|c|c|c|}
\hline
\textbf{Dataset}       & \textbf{With B-Pseudo} & \textbf{With B-VGG} & \textbf{ With B-ResNet} & \textbf{With B-ViT} \\ \hline
\textbf{Cifar10}       & 1.00 (L)& 0.99 (L)& 1.00 (L)& 0.96 (L)\\ \hline
\textbf{Fashion MNIST} & 1.00 (L)& 1.00 (L)& 1.00 (L)& 1.00 (L)\\ \hline
\textbf{MNIST}         & 0.82 (L)& 0.74 (L)& 0.83 (L)& 0.85 (L)\\ \hline
\textbf{SVHN}          & 0.98 (L)& 0.98 (L)& 0.95 (L)& 0.98 (L)\\ \hline
\textbf{CelebA-Hair}  & 1.00 (L)& 1.00 (L)& 1.00 (L)& 1.00 (L)\\ \hline
\textbf{CelebA-M/F
}          & 1.00 (L)& 1.00 (L)& 1.00 (L)& 0.98 (L)\\ \hline
\textbf{Synthetic-Pub1}     & 0.87 (L)& 0.91 (L)& 0.98 (L)& 0.92 (L)\\ \hline
\textbf{Synthetic-Pub2}    & 0.86 (L)& 0.86 (L)& 0.97 (L)& 0.84 (L)\\ \hline
\textbf{Industry}   & 0.93 (L)& 0.93 (L)& 0.97 (L)& 0.91 (L)\\ \hline
\end{tabular}
}
\end{table}

As shown in Figure~\ref{fig:rq2_boxplot}, no single baseline consistently outperforms the others across all datasets. For instance, \textit{B-VGG} is the top-performer among the baselines for CIFAR-10, Fashion MNIST, and MNIST. However, it falls short compared to \textit{B-Pseudo} and \textit{B-ResNet} on other datasets. In contrast, \approach\ surpasses all baselines across all datasets. Specifically, \approach\ yields average accuracy improvements of 
4.6\%, 
7.3\%, 
4.3\% and 
4.5\%
over \textit{B-VGG}, \textit{B-ResNet}, \textit{B-ViT}, and \textit{B-Pseudo}, respectively, leading to an overall average improvement of 4.5\% for all baselines and datasets.

Table~\ref{table:rq2_time} shows the average execution times for each step of \approach\ and the baselines across all datasets.  The data-splitting step has nearly identical execution times across the methods due to its identical implementation. \approach\ requires more time for fine-tuning than the supervised baselines (\textit{B-VGG}, \textit{B-ResNet}, and \textit{B-ViT}) as it fine-tunes three models instead of one. In contrast, \textit{B-Pseudo} is the most time-consuming among all methods, given its costly iterative fine-tuning process over the entire dataset. Finally, the MILP solver is very fast -- significantly faster than fine-tuning, data splitting, and labelling.  Since \approach\ is executed offline rather than within a human feedback loop, its average runtime of approximately 15 minutes does not impede its practical applicability.

\begin{table}[t]
\centering
\caption{Execution time of different steps for \approach\ and baselines averaged over all datasets}
\label{table:rq2_time}
\scalebox{0.83}{
\begin{tabular}{|l|ccccc|}
\hline
\multirow{2}{*}{\textbf{Method}} & \multicolumn{5}{c|}{\textbf{Execution Time (Seconds)}}                                                                                           \\ \cline{2-6} 
                                 & \multicolumn{1}{c|}{\textbf{Data Splitting}} & \multicolumn{1}{c|}{\textbf{Fine-Tuning}} & \multicolumn{1}{c|}{\textbf{MILP Solver}} & \multicolumn{1}{c|}{\textbf{Labelling}} & \textbf{Total} \\ \hline
\textbf{\approach}               & \multicolumn{1}{c|}{
35.1 (4\%)
}               & \multicolumn{1}{c|}{
800.9 (86\%)
}                   & \multicolumn{1}{c|}{
23.8 (3\%)
}            & \multicolumn{1}{c|}{
66.2 (7\%) 
}           & 
926.7
\\ \hline
\textbf{B-VGG}                   & \multicolumn{1}{c|}{39.9 (11\%)}              & \multicolumn{1}{c|}{281.3 (78\%)}                   & \multicolumn{1}{c|}{N/A}                  & \multicolumn{1}{c|}{38.5 (11\%)}          & 359.7         \\ \hline
\textbf{B-ResNet}                & \multicolumn{1}{c|}{40.2 (12\%)}              & \multicolumn{1}{c|}{247.8 (77\%)}                   & \multicolumn{1}{c|}{N/A}                  & \multicolumn{1}{c|}{35.6 (11\%)}          & 323.6         \\ \hline
\textbf{B-ViT}                   & \multicolumn{1}{c|}{43.0 (11\%)}              & \multicolumn{1}{c|}{307.2 (78\%)}                   & \multicolumn{1}{c|}{N/A}                  & \multicolumn{1}{c|}{42.7 (11\%)}          & 392.9         \\ \hline
\textbf{B-Pseudo}                & \multicolumn{1}{c|}{38.7 (2\%)}               & \multicolumn{1}{c|}{2503.8 (98\%)}                 & \multicolumn{1}{c|}{N/A}                  & \multicolumn{1}{c|}{N/A}                 & 2542.5        \\ \hline
\end{tabular}
}

\end{table}




\begin{tcolorbox}[breakable, colback=gray!10!white,colframe=black!75!black]

\textbf{Finding:} Across all 
nine
datasets, and within the same manual labelling budget, \approach\ achieves significantly higher labelling accuracy than all baseline methods, with large effect sizes.

\textbf{Takeway:} Under the same manual labelling effort required by \approach\ in RQ1 to achieve an average near-perfect accuracy of 98.8\%, none of the baselines can exceed an average accuracy of 
94.5\%. Further, \approach's execution time is  feasible given its offline execution model.

\end{tcolorbox}

\textbf{RQ3 experiments.} To answer RQ3, we first apply \approach\ as presented in Algorithm~\ref{alg:optimzationlabel} to the three test-input validation datasets: \textsc{Synthetic-Pub1}, \textsc{Synthetic-Pub2}, and \textsc{Industry}, and compare \approach\ with the four test-input validation baselines described in Section~\ref{subsec:baselines-test-input-validation}. We then extend \approach\ with an active-learning loop and repeat the same experiments. Below, we discuss experiments that compare \approach\ and its active-learning variant against the test-input \hbox{validation baselines.}

\emph{Comparing \approach\ with the test-input validation baselines in Section~\ref{subsec:baselines-test-input-validation}.} For each of our four baselines for test-input validation, we experiment with three manual effort levels: 25\%, 50\% and 75\%. More specifically, for B-VAE, B-VIF and B-HiL-TV1 when experimenting with $x$\% manual effort, we initially train the baseline on a randomly selected $x\%$ of the input dataset 
referred to as the pre-validated subset.
To train B-VAE and B-VIF, we
compute VAE reconstruction error values and VIF values, respectively, for the image pairs in the pre-validated subset.
Then, for each of B-VAE and B-VIF, we vary its thresholds over the range of its respective metric in increments of $10^{-3}$
and calculate the accuracy as the percentage of correctly validated pairs in the pre-validated subset at each threshold value. The threshold leading
to the highest accuracy is selected as the optimal threshold for the baseline.
We then use each of the B-VAE, B-VIF and B-HiL-TV1 to predict validation labels for the remaining $1-x$\% of the dataset.
For B-HiL-TV2, as suggested by the authors of that baseline~\cite{Ghobari_2025_active_learning}, since it uses an active-learning loop, to experiment with $x\%$ manual effort, we initially train the baseline on $\frac{x}{2}\%$ of randomly selected data and then start the active-learning loop. Once the manual effort has reached $x\%$, we stop prompting the human in the loop and label the validity of the rest of the data automatically.
We measure the accuracy and manual effort of each baseline similar to RQ1 and RQ2. 
To address randomness, like in RQ1, we repeat each experiment five times. In total, for the baselines, we perform $4 (\textit{baselines}) \times 3 (\textit{labelling budgets}) \times 3 (\textit{datasets}) \times 5 (\textit{repeats}) = 180$ experiments.

To run \approach\ for test-input validation, we compute the 13 image-comparison metrics from Ghobari et al.~\cite{Ghobari_2025_active_learning} on each original--synthesized image pair. The resulting metrics produce a tabular dataset in which each row represents one image pair. Since DL models are generally ill-suited to tabular data~\cite{shwartz-ziv2021tabular,grinsztajn2022trees}, Algorithm~\ref{alg:optimzationlabel} uses a random-forest classifier, in line with Ghobari et al.'s finding that random forest outperforms other common classification techniques such as SVM, logistic regression, and decision trees for test-input validation.

For each baseline and effort level of 25\%, 50\%, and 75\%, we apply \approach\ to the \textsc{Synthetic-Pub1}, \textsc{Synthetic-Pub2}, and \textsc{Industry} datasets by setting the parameter $\alpha$ in Algorithm~\ref{alg:optimzationlabel} to the average accuracy achieved by five runs of the baseline for that effort level and dataset. By doing so, we investigate how much effort \approach\ requires to match or exceed the accuracies that baselines achieve for different effort levels. This approach allows us to experiment with and compare \approach\ to baselines by setting $\alpha$ to a range of values, and to quantify how much manual effort \approach\ saves compared to each baseline while matching or exceeding the baseline's accuracy.

\emph{Comparing the active-learning variant of \approach\ \hbox{(\approach-AL)} with the test-input validation baselines.} 
Algorithm~\ref{alg:milp_active_learning} presents a variant of \approach, referred to as \approach-AL, which integrates active learning. The inputs and parameters of \approach-AL are identical to those of Algorithm~\ref{alg:optimzationlabel}, except for the additional parameter $\beta$, which denotes the number of samples manually validated in each iteration of the active-learning loop. Similar to Algorithm~\ref{alg:optimzationlabel},  \approach-AL starts by performing the data splitting routine (line~1 of Algorithm~\ref{alg:milp_active_learning}) and manually validating the fine-tuning dataset, $D_t$, and the optimization dataset, $D_o$ (line~2 of Algorithm~\ref{alg:milp_active_learning}). Each iteration of the active-learning loop (lines 3--12  of Algorithm~\ref{alg:milp_active_learning})  starts by executing lines 3--16 of Algorithm~\ref{alg:optimzationlabel} on sets $D_t$, $D_o$ and $D'$, which is the to-be-labelled dataset. By doing so, \approach's MILP formulation automatically validates a subset of $D'$ and identifies a susbet $D_m$ of $D'$ that cannot be automatically validated with high accuracy.  Instead of requesting  manual validation for the entire $D_m$ as done in \approach, \approach-AL randomly selects a subset of $D_m$ with at most $\beta$ elements and requests the human validator  to validate this smaller subset (lines 5--7 of Algorithm~\ref{alg:milp_active_learning}). Then the sets $D'$, $D_t$ and $D_o$ are updated, by removing the manually validated elements from $D'$, and adding them to $D_t$ and $D_o$ (lines~8--11 of Algorithm~\ref{alg:milp_active_learning}).
The active-learning loop of \approach-AL continues until all of the elements in $D'$ have been either automatically or manually validated.

For the parameters of \approach-AL, we use the same configuration as that applied for comparing \approach\ with the test-input validation baselines. When comparing \approach-AL with \approach, the only new parameter is $\beta$. Ghobari et al.~\cite{Ghobari_2025_active_learning} showed that $\beta$ affects only the number of iterations and has no significant impact on the accuracy or manual effort of test-input validation. 
We investigated \approach-AL with $\beta$ values of 10, 50, 100, and 150, which confirmed this observation. In the experimental results reported for RQ3, we use $\beta = 50$.

\small
\begin{algorithm}[t]
\caption{\approach\ with Active Learning (\approach-AL)}
\label{alg:milp_active_learning}
{\footnotesize
\begin{flushleft}
\textbf{Input} $\mathit{D}$: An unlabelled dataset \\ 
\textbf{Input} $\alpha$: Accuracy threshold \\
\textbf{Input} $c_1, \ldots, c_n$: A set of pretrained classifiers\\
\textbf{Param} $\mathit{hInitial}, s$: \ldots \\
\textbf{Param} $\beta$: Maximum human validations per active-learning iteration \CommentG{\textbf{The inputs, parameters, and output of OPAL-AL are the same as those of 
OPAL (Algorithm~\ref{alg:optimzationlabel}), except for the new $\beta$ parameter.}}\\[.3em]
\textbf{Output} $D$: Labelled dataset\\
\end{flushleft}
\begin{algorithmic}[1]

\State $(D_t,  D_o, D')=\text{DataSplit}(D, \mathit{hInitial}, s)$ \CommentG{\textbf{Line 1 of Algorithm~\ref{alg:optimzationlabel}}}
\State Manually validate $D_t \cup D_o$ \CommentG{\textbf{Line 2 of Algorithm~\ref{alg:optimzationlabel}}}
\While{$D' \neq \emptyset$}

    \State  Execute lines 3 to 16 of \approach\  (Algorithm~\ref{alg:optimzationlabel}) 
  on $D_t$, $D_o$, and $D'$ 
    \State $D' \gets D_m$ \CommentG{\textbf{$D_m$ is the set of data to be manually validated as computed by the loop from lines 5--16 of Algorithm~\ref{alg:optimzationlabel}}}

    \State Randomly select $D_{\mathit{tmp}}\subseteq D'$ with $|D_{\mathit{tmp}}|=\min(\beta,|D'|)$
    \State Prompt human to validate all pairs in $D_{\mathit{tmp}}$
    \State $D'\gets D'\setminus D_{\mathit{tmp}}$
    \State Partition $D_{\mathit{tmp}}$ into two equal random splits $D_{\mathit{tmp},1}$ and $D_{\mathit{tmp},2}$. 
      \State $D_{t}\gets D_t \cup D_{\mathit{tmp},1}$  
      
      \State $D_{o}\gets D_o \cup D_{\mathit{tmp},2}$ 
      
\EndWhile
\State \Return $D$
\end{algorithmic}}
\end{algorithm}
\normalsize

\begin{table}[t]
\centering
\caption{Comparing \approach\ and \approach\ extended with active learning (i.e., \approach-AL) with  four test-input validation baselines described in Section~\ref{subsec:baselines-test-input-validation} for \textsc{Synthetic-Pub1} dataset.
For comparisons yielding a significance level of \(p < 0.01\), the corresponding values of \(\Delta_{\mathrm{ME}}\) and \(\Delta_{\mathrm{Acc}}\) are emphasized in \textbf{boldface}. For manual effort and accuracy, if \approach\ or \approach-AL has outperformed a baseline, the corresponding $\Delta$ value is highlighted in green. 
}
\label{table:rq3_baselines_synthetic1}
\scalebox{0.85}{
\begin{tabular}{|cc|cc|cc|cc|cc|}
\hline
\rowcolor[HTML]{C0C0C0} 
\multicolumn{2}{|c|}{\cellcolor[HTML]{C0C0C0}B-HiL-TV2}                                                 & \multicolumn{2}{c|}{\cellcolor[HTML]{C0C0C0}OPAL}                                                   & \multicolumn{2}{c|}{\cellcolor[HTML]{C0C0C0}OPAL vs B-HiL-TV2}                                                          & \multicolumn{2}{c|}{\cellcolor[HTML]{C0C0C0}OPAL-AL}                                                & \multicolumn{2}{c|}{\cellcolor[HTML]{C0C0C0}OPAL-AL vs B-HiL-TV2}                                                       \\ \hline
\rowcolor[HTML]{C0C0C0} 
\multicolumn{1}{|c|}{\cellcolor[HTML]{C0C0C0}ME}                        & Acc                        & \multicolumn{1}{c|}{\cellcolor[HTML]{C0C0C0}ME}                        & Acc                        & \multicolumn{1}{c|}{\cellcolor[HTML]{C0C0C0}$\Delta_{ME}$}                        & $\Delta_{Acc}$                        & \multicolumn{1}{c|}{\cellcolor[HTML]{C0C0C0}ME}                        & Acc                        & \multicolumn{1}{c|}{\cellcolor[HTML]{C0C0C0}$\Delta_{ME}$}                        & $\Delta_{Acc}$                        \\ \hline
\multicolumn{1}{|c|}{0.25}                                              & 0.91                       & \multicolumn{1}{c|}{0.09}                                              & 0.91                       & \multicolumn{1}{c|}{\cellcolor{green!20}\textbf{-0.16}}                                              & \cellcolor{green!20}0                        & \multicolumn{1}{c|}{0.18}                                              & 0.92                       & \multicolumn{1}{c|}{\cellcolor{green!20}\textbf{-0.07}}                                              & \cellcolor{green!20}0.01                     \\ \hline
\multicolumn{1}{|c|}{0.50}                                              & 0.96                       & \multicolumn{1}{c|}{0.65}                                              & 0.97                       & \multicolumn{1}{c|}{0.15}                                                        & \cellcolor{green!20}0.01                     & \multicolumn{1}{c|}{0.46}                                              & 0.96                       & \multicolumn{1}{c|}{\cellcolor{green!20}\textbf{-0.04}}                                              & \cellcolor{green!20}0                        \\ \hline
\multicolumn{1}{|c|}{0.75}                                              & 0.99                       & \multicolumn{1}{c|}{0.81}                                              & 0.99                       & \multicolumn{1}{c|}{0.06}                                                        & \cellcolor{green!20}0                        & \multicolumn{1}{c|}{0.56}                                              & 0.97                       & \multicolumn{1}{c|}{\cellcolor{green!20}\textbf{-0.19}}                                              & -0.02                             \\ \hline
\rowcolor[HTML]{C0C0C0} 
\multicolumn{2}{|c|}{\cellcolor[HTML]{C0C0C0}{\color[HTML]{000000} B-HiL-TV1}}                   & \multicolumn{2}{c|}{\cellcolor[HTML]{C0C0C0}{\color[HTML]{000000} OPAL}}                            & \multicolumn{2}{c|}{\cellcolor[HTML]{C0C0C0}{\color[HTML]{000000} OPAL vs B-HiL-TV1}}                            & \multicolumn{2}{c|}{\cellcolor[HTML]{C0C0C0}{\color[HTML]{000000} OPAL-AL}}                         & \multicolumn{2}{c|}{\cellcolor[HTML]{C0C0C0}{\color[HTML]{000000} OPAL-AL vs B-HiL-TV1}}                         \\ \hline
\rowcolor[HTML]{C0C0C0} 
\multicolumn{1}{|c|}{\cellcolor[HTML]{C0C0C0}{\color[HTML]{000000} ME}} & {\color[HTML]{000000} Acc} & \multicolumn{1}{c|}{\cellcolor[HTML]{C0C0C0}{\color[HTML]{000000} ME}} & {\color[HTML]{000000} Acc} & \multicolumn{1}{c|}{\cellcolor[HTML]{C0C0C0}{\color[HTML]{000000} $\Delta_{ME}$}} & {\color[HTML]{000000} $\Delta_{Acc}$} & \multicolumn{1}{c|}{\cellcolor[HTML]{C0C0C0}{\color[HTML]{000000} ME}} & {\color[HTML]{000000} Acc} & \multicolumn{1}{c|}{\cellcolor[HTML]{C0C0C0}{\color[HTML]{000000} $\Delta_{ME}$}} & {\color[HTML]{000000} $\Delta_{Acc}$} \\ \hline
\multicolumn{1}{|c|}{0.25}                                              & 0.86                       & \multicolumn{1}{c|}{0.09}                                              & 0.9                        & \multicolumn{1}{c|}{\cellcolor{green!20}\textbf{-0.16}}                                              & \cellcolor{green!20}\textbf{0.04}                     & \multicolumn{1}{c|}{0.07}                                              & 0.9                        & \multicolumn{1}{c|}{\cellcolor{green!20}\textbf{-0.18}}                                              & \cellcolor{green!20}\textbf{0.04}                     \\ \hline
\multicolumn{1}{|c|}{0.50}                                              & 0.87                       & \multicolumn{1}{c|}{0.09}                                              & 0.9                        & \multicolumn{1}{c|}{\cellcolor{green!20}\textbf{-0.41}}                                              & \cellcolor{green!20}\textbf{0.03}                     & \multicolumn{1}{c|}{0.07}                                              & 0.9                        & \multicolumn{1}{c|}{\cellcolor{green!20}\textbf{-0.43}}                                              & \cellcolor{green!20}\textbf{0.03}                     \\ \hline
\multicolumn{1}{|c|}{0.75}                                              & 0.88                       & \multicolumn{1}{c|}{0.07}                                              & 0.89                       & \multicolumn{1}{c|}{\cellcolor{green!20}\textbf{-0.68}}                                              & \cellcolor{green!20}0.01                     & \multicolumn{1}{c|}{0.07}                                              & 0.9                        & \multicolumn{1}{c|}{\cellcolor{green!20}\textbf{-0.68}}                                              & \cellcolor{green!20}\textbf{0.02}                     \\ \hline
\rowcolor[HTML]{C0C0C0} 
\multicolumn{2}{|c|}{\cellcolor[HTML]{C0C0C0}B-VIF}                                                    & \multicolumn{2}{c|}{\cellcolor[HTML]{C0C0C0}OPAL}                                                   & \multicolumn{2}{c|}{\cellcolor[HTML]{C0C0C0}OPAL vs B-VIF}                                                             & \multicolumn{2}{c|}{\cellcolor[HTML]{C0C0C0}OPAL-AL}                                                & \multicolumn{2}{c|}{\cellcolor[HTML]{C0C0C0}OPAL-AL vs B-VIF}                                                          \\ \hline
\rowcolor[HTML]{C0C0C0} 
\multicolumn{1}{|c|}{\cellcolor[HTML]{C0C0C0}ME}                        & Acc                        & \multicolumn{1}{c|}{\cellcolor[HTML]{C0C0C0}ME}                        & Acc                        & \multicolumn{1}{c|}{\cellcolor[HTML]{C0C0C0}$\Delta_{ME}$}                        & $\Delta_{Acc}$                        & \multicolumn{1}{c|}{\cellcolor[HTML]{C0C0C0}ME}                        & Acc                        & \multicolumn{1}{c|}{\cellcolor[HTML]{C0C0C0}$\Delta_{ME}$}                        & $\Delta_{Acc}$                        \\ \hline
\multicolumn{1}{|c|}{0.25}                                              & 0.81                       & \multicolumn{1}{c|}{0.08}                                              & 0.89                       & \multicolumn{1}{c|}{\cellcolor{green!20}\textbf{-0.17}}                                              & \cellcolor{green!20}\textbf{0.08}                     & \multicolumn{1}{c|}{0.06}                                              & 0.89                       & \multicolumn{1}{c|}{\cellcolor{green!20}\textbf{-0.19}}                                              & \cellcolor{green!20}\textbf{0.08}                     \\ \hline
\multicolumn{1}{|c|}{0.50}                                              & 0.81                       & \multicolumn{1}{c|}{0.08}                                              & 0.89                       & \multicolumn{1}{c|}{\cellcolor{green!20}\textbf{-0.42}}                                              & \cellcolor{green!20}\textbf{0.08}                     & \multicolumn{1}{c|}{0.06}                                              & 0.89                       & \multicolumn{1}{c|}{\cellcolor{green!20}\textbf{-0.44}}                                              & \cellcolor{green!20}\textbf{0.08}                     \\ \hline
\multicolumn{1}{|c|}{0.75}                                              & 0.81                       & \multicolumn{1}{c|}{0.08}                                              & 0.89                       & \multicolumn{1}{c|}{\cellcolor{green!20}\textbf{-0.67}}                                              & \cellcolor{green!20}\textbf{0.08}                     & \multicolumn{1}{c|}{0.06}                                              & 0.89                       & \multicolumn{1}{c|}{\cellcolor{green!20}\textbf{-0.69}}                                              & \cellcolor{green!20}\textbf{0.08}                     \\ \hline
\rowcolor[HTML]{C0C0C0} 
\multicolumn{2}{|c|}{\cellcolor[HTML]{C0C0C0}B-VAE}                                                    & \multicolumn{2}{c|}{\cellcolor[HTML]{C0C0C0}OPAL}                                                   & \multicolumn{2}{c|}{\cellcolor[HTML]{C0C0C0}OPAL vs B-VAE}                                                             & \multicolumn{2}{c|}{\cellcolor[HTML]{C0C0C0}OPAL-AL}                                                & \multicolumn{2}{c|}{\cellcolor[HTML]{C0C0C0}OPAL-AL vs B-VAE}                                                          \\ \hline
\rowcolor[HTML]{C0C0C0} 
\multicolumn{1}{|c|}{\cellcolor[HTML]{C0C0C0}ME}                        & Acc                        & \multicolumn{1}{c|}{\cellcolor[HTML]{C0C0C0}ME}                        & Acc                        & \multicolumn{1}{c|}{\cellcolor[HTML]{C0C0C0}$\Delta_{ME}$}                        & $\Delta_{Acc}$                        & \multicolumn{1}{c|}{\cellcolor[HTML]{C0C0C0}ME}                        & Acc                        & \multicolumn{1}{c|}{\cellcolor[HTML]{C0C0C0}$\Delta_{ME}$}                        & $\Delta_{Acc}$                        \\ \hline
\multicolumn{1}{|c|}{0.25}                                              & 0.81                       & \multicolumn{1}{c|}{0.08}                                              & 0.89                       & \multicolumn{1}{c|}{\cellcolor{green!20}\textbf{-0.17}}                                              & \cellcolor{green!20}\textbf{0.08}                     & \multicolumn{1}{c|}{0.06}                                              & 0.89                       & \multicolumn{1}{c|}{\cellcolor{green!20}\textbf{-0.19}}                                              & \cellcolor{green!20}\textbf{0.08}                     \\ \hline
\multicolumn{1}{|c|}{0.50}                                              & 0.81                       & \multicolumn{1}{c|}{0.08}                                              & 0.89                       & \multicolumn{1}{c|}{\cellcolor{green!20}\textbf{-0.42}}                                              & \cellcolor{green!20}\textbf{0.08}                     & \multicolumn{1}{c|}{0.06}                                              & 0.89                       & \multicolumn{1}{c|}{\cellcolor{green!20}\textbf{-0.44}}                                              & \cellcolor{green!20}\textbf{0.08}                     \\ \hline
\multicolumn{1}{|c|}{0.75}                                              & 0.81                       & \multicolumn{1}{c|}{0.08}                                              & 0.89                       & \multicolumn{1}{c|}{\cellcolor{green!20}\textbf{-0.67}}                                              & \cellcolor{green!20}\textbf{0.08}                     & \multicolumn{1}{c|}{0.06}                                              & 0.89                       & \multicolumn{1}{c|}{\cellcolor{green!20}\textbf{-0.69}}                                              & \cellcolor{green!20}\textbf{0.08}                     \\ \hline
\rowcolor[HTML]{C0C0C0} 
\multicolumn{2}{|c|}{\cellcolor[HTML]{C0C0C0}Average}                                                & \multicolumn{1}{c|}{\cellcolor[HTML]{C0C0C0}}                          &                            & \multicolumn{1}{c|}{\cellcolor{green!20}\textbf{-0.31 (L)}}                               & \cellcolor{green!20}\textbf{0.048 (M)}                            & \multicolumn{1}{c|}{\cellcolor[HTML]{C0C0C0}}                          &                            & \multicolumn{1}{c|}{\cellcolor{green!20}\textbf{-0.352 (L)}}                              & \cellcolor{green!20}\textbf{0.047 (M)}                             \\ \hline
\end{tabular}

}
\end{table}

\begin{table*}[t]
\centering
\caption{Comparing \approach\ and \approach\ extended with active learning (i.e., \approach-AL) with  four test-input validation baselines described in Section~\ref{subsec:baselines-test-input-validation} for \textsc{Synthetic-Pub2} dataset
For comparisons yielding a significance level of \(p < 0.01\), the corresponding values of \(\Delta_{\mathrm{ME}}\) and \(\Delta_{\mathrm{Acc}}\) are emphasized in \textbf{boldface}. For manual effort and accuracy, if \approach\ or \approach-AL has outperformed a baseline, the corresponding $\Delta$ value is highlighted in green.
}
\label{table:rq3_baselines_synthetic2}
\scalebox{0.85}{
\begin{tabular}{|cc|cc|cc|cc|cc|}
\hline
\rowcolor[HTML]{C0C0C0} 
\multicolumn{2}{|c|}{\cellcolor[HTML]{C0C0C0}B-HiL-TV2}        & \multicolumn{2}{c|}{\cellcolor[HTML]{C0C0C0}OPAL}      & \multicolumn{2}{c|}{\cellcolor[HTML]{C0C0C0}OPAL vs B-HiL-TV2}                  & \multicolumn{2}{c|}{\cellcolor[HTML]{C0C0C0}OPAL-AL}   & \multicolumn{2}{c|}{\cellcolor[HTML]{C0C0C0}OPAL-AL vs B-HiL-TV2}              \\ \hline
\rowcolor[HTML]{C0C0C0} 
\multicolumn{1}{|c|}{\cellcolor[HTML]{C0C0C0}ME}   & Acc    & \multicolumn{1}{c|}{\cellcolor[HTML]{C0C0C0}ME} & Acc  & \multicolumn{1}{c|}{\cellcolor[HTML]{C0C0C0}$\Delta_{ME}$}   & $\Delta_{Acc}$     & \multicolumn{1}{c|}{\cellcolor[HTML]{C0C0C0}ME} & Acc  & \multicolumn{1}{c|}{\cellcolor[HTML]{C0C0C0}$\Delta_{ME}$}   & $\Delta_{Acc}$    \\ \hline
\multicolumn{1}{|c|}{0.25}                         & 0.9    & \multicolumn{1}{c|}{0.19}                       & 0.91 & \multicolumn{1}{c|}{\cellcolor{green!20}\textbf{-0.06}}      & \cellcolor{green!20}0.01  & \multicolumn{1}{c|}{0.17}                       & 0.91 & \multicolumn{1}{c|}{\cellcolor{green!20}\textbf{-0.08}}      & \cellcolor{green!20}\textbf{0.01} \\ \hline
\multicolumn{1}{|c|}{0.50}                         & 0.92   & \multicolumn{1}{c|}{0.26}                       & 0.94 & \multicolumn{1}{c|}{\cellcolor{green!20}\textbf{-0.24}}      & \cellcolor{green!20}\textbf{0.02}  & \multicolumn{1}{c|}{0.25}                       & 0.94 & \multicolumn{1}{c|}{\cellcolor{green!20}\textbf{-0.25}}      & \cellcolor{green!20}\textbf{0.02} \\ \hline
\multicolumn{1}{|c|}{0.75}                         & 0.97   & \multicolumn{1}{c|}{0.42}                       & 0.97 & \multicolumn{1}{c|}{\cellcolor{green!20}\textbf{-0.33}}      & \cellcolor{green!20}0     & \multicolumn{1}{c|}{0.39}                       & 0.97 & \multicolumn{1}{c|}{\cellcolor{green!20}\textbf{-0.36}}      & \cellcolor{green!20}0    \\ \hline
\rowcolor[HTML]{C0C0C0} 
\multicolumn{2}{|c|}{\cellcolor[HTML]{C0C0C0}B-HiL-TV1} & \multicolumn{2}{c|}{\cellcolor[HTML]{C0C0C0}OPAL}      & \multicolumn{2}{c|}{\cellcolor[HTML]{C0C0C0}OPAL vs B-HiL-TV1}           & \multicolumn{2}{c|}{\cellcolor[HTML]{C0C0C0}OPAL-AL}   & \multicolumn{2}{c|}{\cellcolor[HTML]{C0C0C0}OPAL-AL vs B-HiL-TV1}       \\ \hline
\rowcolor[HTML]{C0C0C0} 
\multicolumn{1}{|c|}{\cellcolor[HTML]{C0C0C0}ME}   & Acc    & \multicolumn{1}{c|}{\cellcolor[HTML]{C0C0C0}ME} & Acc  & \multicolumn{1}{c|}{\cellcolor[HTML]{C0C0C0}$\Delta_{ME}$}   & $\Delta_{Acc}$     & \multicolumn{1}{c|}{\cellcolor[HTML]{C0C0C0}ME} & Acc  & \multicolumn{1}{c|}{\cellcolor[HTML]{C0C0C0}$\Delta_{ME}$}   & $\Delta_{Acc}$    \\ \hline
\multicolumn{1}{|c|}{0.25}                         & 0.88   & \multicolumn{1}{c|}{0.14}                       & 0.88 & \multicolumn{1}{c|}{\cellcolor{green!20}\textbf{-0.11}}      & \cellcolor{green!20}0     & \multicolumn{1}{c|}{0.11}                       & 0.88 & \multicolumn{1}{c|}{\cellcolor{green!20}\textbf{-0.14}}      & \cellcolor{green!20}0    \\ \hline
\multicolumn{1}{|c|}{0.50}                         & 0.89   & \multicolumn{1}{c|}{0.24}                       & 0.93 & \multicolumn{1}{c|}{\cellcolor{green!20}\textbf{-0.26}}      & \cellcolor{green!20}\textbf{0.04}  & \multicolumn{1}{c|}{0.21}                       & 0.93 & \multicolumn{1}{c|}{\cellcolor{green!20}\textbf{-0.29}}      & \cellcolor{green!20}\textbf{0.04} \\ \hline
\multicolumn{1}{|c|}{0.75}                         & 0.96   & \multicolumn{1}{c|}{0.39}                       & 0.96 & \multicolumn{1}{c|}{\cellcolor{green!20}\textbf{-0.36}}      & \cellcolor{green!20}0     & \multicolumn{1}{c|}{0.38}                       & 0.96 & \multicolumn{1}{c|}{\cellcolor{green!20}\textbf{-0.37}}      & \cellcolor{green!20}0    \\ \hline
\rowcolor[HTML]{C0C0C0} 
\multicolumn{2}{|c|}{\cellcolor[HTML]{C0C0C0}B-VIF}           & \multicolumn{2}{c|}{\cellcolor[HTML]{C0C0C0}OPAL}      & \multicolumn{2}{c|}{\cellcolor[HTML]{C0C0C0}OPAL vs B-VIF}                     & \multicolumn{2}{c|}{\cellcolor[HTML]{C0C0C0}OPAL-AL}   & \multicolumn{2}{c|}{\cellcolor[HTML]{C0C0C0}OPAL-AL vs B-VIF}                 \\ \hline
\rowcolor[HTML]{C0C0C0} 
\multicolumn{1}{|c|}{\cellcolor[HTML]{C0C0C0}ME}   & Acc    & \multicolumn{1}{c|}{\cellcolor[HTML]{C0C0C0}ME} & Acc  & \multicolumn{1}{c|}{\cellcolor[HTML]{C0C0C0}$\Delta_{ME}$}   & $\Delta_{Acc}$     & \multicolumn{1}{c|}{\cellcolor[HTML]{C0C0C0}ME} & Acc  & \multicolumn{1}{c|}{\cellcolor[HTML]{C0C0C0}$\Delta_{ME}$}   & $\Delta_{Acc}$    \\ \hline
\multicolumn{1}{|c|}{0.25}                         & 0.86   & \multicolumn{1}{c|}{0.13}                       & 0.87 & \multicolumn{1}{c|}{\cellcolor{green!20}\textbf{-0.12}}      & \cellcolor{green!20}0.01  & \multicolumn{1}{c|}{0.1}                        & 0.86 & \multicolumn{1}{c|}{\cellcolor{green!20}\textbf{-0.15}}      & \cellcolor{green!20}0    \\ \hline
\multicolumn{1}{|c|}{0.50}                         & 0.86   & \multicolumn{1}{c|}{0.24}                       & 0.93 & \multicolumn{1}{c|}{\cellcolor{green!20}\textbf{-0.26}}      & \cellcolor{green!20}\textbf{0.07}  & \multicolumn{1}{c|}{0.1}                        & 0.93 & \multicolumn{1}{c|}{\cellcolor{green!20}\textbf{-0.4}}       & \cellcolor{green!20}\textbf{0.07} \\ \hline
\multicolumn{1}{|c|}{0.75}                         & 0.95   & \multicolumn{1}{c|}{0.39}                       & 0.95 & \multicolumn{1}{c|}{\cellcolor{green!20}\textbf{-0.36}}      & \cellcolor{green!20}0     & \multicolumn{1}{c|}{0.36}                       & 0.95 & \multicolumn{1}{c|}{\cellcolor{green!20}\textbf{-0.39}}      & \cellcolor{green!20}0    \\ \hline
\rowcolor[HTML]{C0C0C0} 
\multicolumn{2}{|c|}{\cellcolor[HTML]{C0C0C0}B-VAE}           & \multicolumn{2}{c|}{\cellcolor[HTML]{C0C0C0}OPAL}      & \multicolumn{2}{c|}{\cellcolor[HTML]{C0C0C0}OPAL vs B-VAE}                     & \multicolumn{2}{c|}{\cellcolor[HTML]{C0C0C0}OPAL-AL}   & \multicolumn{2}{c|}{\cellcolor[HTML]{C0C0C0}OPAL-AL vs B-VAE}                 \\ \hline
\rowcolor[HTML]{C0C0C0} 
\multicolumn{1}{|c|}{\cellcolor[HTML]{C0C0C0}ME}   & Acc    & \multicolumn{1}{c|}{\cellcolor[HTML]{C0C0C0}ME} & Acc  & \multicolumn{1}{c|}{\cellcolor[HTML]{C0C0C0}$\Delta_{ME}$}   & $\Delta_{Acc}$     & \multicolumn{1}{c|}{\cellcolor[HTML]{C0C0C0}ME} & Acc  & \multicolumn{1}{c|}{\cellcolor[HTML]{C0C0C0}$\Delta_{ME}$}   & $\Delta_{Acc}$    \\ \hline
\multicolumn{1}{|c|}{0.25}                         & 0.82   & \multicolumn{1}{c|}{0.11}                       & 0.87 & \multicolumn{1}{c|}{\cellcolor{green!20}\textbf{-0.14}}      & \cellcolor{green!20}\textbf{0.05}  & \multicolumn{1}{c|}{0.1}                        & 0.85 & \multicolumn{1}{c|}{\cellcolor{green!20}\textbf{-0.15}}      & \cellcolor{green!20}\textbf{0.03} \\ \hline
\multicolumn{1}{|c|}{0.50}                         & 0.83   & \multicolumn{1}{c|}{0.24}                       & 0.93 & \multicolumn{1}{c|}{\cellcolor{green!20}\textbf{-0.26}}      & \cellcolor{green!20}\textbf{0.1}   & \multicolumn{1}{c|}{0.1}                        & 0.91 & \multicolumn{1}{c|}{\cellcolor{green!20}\textbf{-0.4}}       & \cellcolor{green!20}\textbf{0.08} \\ \hline
\multicolumn{1}{|c|}{0.75}                         & 0.95   & \multicolumn{1}{c|}{0.39}                       & 0.95 & \multicolumn{1}{c|}{\cellcolor{green!20}\textbf{-0.36}}      & \cellcolor{green!20}0     & \multicolumn{1}{c|}{0.36}                       & 0.95 & \multicolumn{1}{c|}{\cellcolor{green!20}\textbf{-0.39}}      & \cellcolor{green!20}0    \\ \hline
\rowcolor[HTML]{C0C0C0} 
\multicolumn{2}{|c|}{\cellcolor[HTML]{C0C0C0}Average}       & \multicolumn{1}{c|}{\cellcolor[HTML]{C0C0C0}}   &      & \multicolumn{1}{c|}{\cellcolor[HTML]{C0C0C0}\cellcolor{green!20}\textbf{-0.24 (L)}} & \cellcolor{green!20}\textbf{0.025 (M)} & \multicolumn{1}{c|}{\cellcolor[HTML]{C0C0C0}}   &      & \multicolumn{1}{c|}{\cellcolor[HTML]{C0C0C0}\cellcolor{green!20}\textbf{-0.28 (L)}} & \cellcolor{green!20}\textbf{0.02 (M)} \\ \hline
\end{tabular}

}
\end{table*}

\begin{table*}[t]
\centering
\caption{Comparing \approach\ and \approach\ extended with active learning (i.e., \approach-AL) with  four test-input validation baselines described in Section~\ref{subsec:baselines-test-input-validation} for  \textsc{Industry} dataset
For comparisons yielding a significance level of \(p < 0.01\), the corresponding values of \(\Delta_{\mathrm{ME}}\) and \(\Delta_{\mathrm{Acc}}\) are emphasized in \textbf{boldface}. For manual effort and accuracy, if \approach\ or \approach-AL has outperformed a baseline, the corresponding $\Delta$ value is highlighted in green.
}
\label{table:rq3_baselines_industry}
\scalebox{0.85}{
\begin{tabular}{|cc|cc|cc|cc|cc|}
\hline
\rowcolor[HTML]{C0C0C0} 
\multicolumn{2}{|c|}{\cellcolor[HTML]{C0C0C0}B-HiL-TV2}         & \multicolumn{2}{c|}{\cellcolor[HTML]{C0C0C0}OPAL}      & \multicolumn{2}{c|}{\cellcolor[HTML]{C0C0C0}OPAL vs B-HiL-TV2}                    & \multicolumn{2}{c|}{\cellcolor[HTML]{C0C0C0}OPAL-AL}   & \multicolumn{2}{c|}{\cellcolor[HTML]{C0C0C0}OPAL-AL vs B-HiL-TV2}               \\ \hline
\rowcolor[HTML]{C0C0C0} 
\multicolumn{1}{|c|}{\cellcolor[HTML]{C0C0C0}ME}      & Acc  & \multicolumn{1}{c|}{\cellcolor[HTML]{C0C0C0}ME} & Acc  & \multicolumn{1}{c|}{\cellcolor[HTML]{C0C0C0}$\Delta_{ME}$}    & $\Delta_{Acc}$      & \multicolumn{1}{c|}{\cellcolor[HTML]{C0C0C0}ME} & Acc  & \multicolumn{1}{c|}{\cellcolor[HTML]{C0C0C0}$\Delta_{ME}$}   & $\Delta_{Acc}$     \\ \hline
\multicolumn{1}{|c|}{0.25} & 0.97 & \multicolumn{1}{c|}{0.32} & 0.97 & \multicolumn{1}{c|}{\textbf{0.07}} & \cellcolor{green!20}0 & \multicolumn{1}{c|}{0.11} & 0.94 & \multicolumn{1}{c|}{\cellcolor{green!20}\textbf{-0.14}} & -0.03 \\ \hline
\multicolumn{1}{|c|}{0.50} & 0.99 & \multicolumn{1}{c|}{0.46} & 0.98 & \multicolumn{1}{c|}{\cellcolor{green!20}\textbf{-0.04}} & \cellcolor{green!20}-0.01 & \multicolumn{1}{c|}{0.33} & 0.97 & \multicolumn{1}{c|}{\cellcolor{green!20}\textbf{-0.17}} & -0.02 \\ \hline
\multicolumn{1}{|c|}{0.75} & 1.00 & \multicolumn{1}{c|}{0.81} & 1    & \multicolumn{1}{c|}{0.06} & \cellcolor{green!20}0 & \multicolumn{1}{c|}{0.63} & 1    & \multicolumn{1}{c|}{\cellcolor{green!20}\textbf{-0.12}} & \cellcolor{green!20}0 \\ \hline
\rowcolor[HTML]{C0C0C0} 
\multicolumn{2}{|c|}{\cellcolor[HTML]{C0C0C0}B-HiL-TV1}  & \multicolumn{2}{c|}{\cellcolor[HTML]{C0C0C0}OPAL}      & \multicolumn{2}{c|}{\cellcolor[HTML]{C0C0C0}OPAL vs B-HiL-TV1}             & \multicolumn{2}{c|}{\cellcolor[HTML]{C0C0C0}OPAL-AL}   & \multicolumn{2}{c|}{\cellcolor[HTML]{C0C0C0}OPAL-AL vs B-HiL-TV1}        \\ \hline
\rowcolor[HTML]{C0C0C0} 
\multicolumn{1}{|c|}{\cellcolor[HTML]{C0C0C0}ME}      & Acc  & \multicolumn{1}{c|}{\cellcolor[HTML]{C0C0C0}ME} & Acc  & \multicolumn{1}{c|}{\cellcolor[HTML]{C0C0C0}$\Delta_{ME}$}    & $\Delta_{Acc}$      & \multicolumn{1}{c|}{\cellcolor[HTML]{C0C0C0}ME} & Acc  & \multicolumn{1}{c|}{\cellcolor[HTML]{C0C0C0}$\Delta_{ME}$}   & $\Delta_{Acc}$     \\ \hline
\multicolumn{1}{|c|}{0.25} & 0.92 & \multicolumn{1}{c|}{0.09} & 0.92 & \multicolumn{1}{c|}{\cellcolor{green!20}\textbf{-0.16}} & \cellcolor{green!20}0 & \multicolumn{1}{c|}{0.06} & 0.91 & \multicolumn{1}{c|}{\cellcolor{green!20}\textbf{-0.19}} & -0.01 \\ \hline
\multicolumn{1}{|c|}{0.50} & 0.92 & \multicolumn{1}{c|}{0.09} & 0.92 & \multicolumn{1}{c|}{\cellcolor{green!20}\textbf{-0.41}} & \cellcolor{green!20}0 & \multicolumn{1}{c|}{0.06} & 0.91 & \multicolumn{1}{c|}{\cellcolor{green!20}\textbf{-0.44}} & -0.01 \\ \hline
\multicolumn{1}{|c|}{0.75} & 0.92 & \multicolumn{1}{c|}{0.09} & 0.92 & \multicolumn{1}{c|}{\cellcolor{green!20}\textbf{-0.66}} & \cellcolor{green!20}0 & \multicolumn{1}{c|}{0.06} & 0.91 & \multicolumn{1}{c|}{\cellcolor{green!20}\textbf{-0.69}} & -0.01 \\ \hline
\rowcolor[HTML]{C0C0C0} 
\multicolumn{2}{|c|}{\cellcolor[HTML]{C0C0C0}B-VIF}            & \multicolumn{2}{c|}{\cellcolor[HTML]{C0C0C0}OPAL}      & \multicolumn{2}{c|}{\cellcolor[HTML]{C0C0C0}OPAL vs B-VIF}                       & \multicolumn{2}{c|}{\cellcolor[HTML]{C0C0C0}OPAL-AL}   & \multicolumn{2}{c|}{\cellcolor[HTML]{C0C0C0}OPAL-AL vs B-VIF}                  \\ \hline
\rowcolor[HTML]{C0C0C0} 
\multicolumn{1}{|c|}{\cellcolor[HTML]{C0C0C0}ME}      & Acc  & \multicolumn{1}{c|}{\cellcolor[HTML]{C0C0C0}ME} & Acc  & \multicolumn{1}{c|}{\cellcolor[HTML]{C0C0C0}$\Delta_{ME}$}    & $\Delta_{Acc}$      & \multicolumn{1}{c|}{\cellcolor[HTML]{C0C0C0}ME} & Acc  & \multicolumn{1}{c|}{\cellcolor[HTML]{C0C0C0}$\Delta_{ME}$}   & $\Delta_{Acc}$     \\ \hline
\multicolumn{1}{|c|}{0.25} & 0.88 & \multicolumn{1}{c|}{0.05} & 0.91 & \multicolumn{1}{c|}{\cellcolor{green!20}\textbf{-0.2}} & \cellcolor{green!20}\textbf{0.03} & \multicolumn{1}{c|}{0.05} & 0.9  & \multicolumn{1}{c|}{\cellcolor{green!20}\textbf{-0.2}} & \cellcolor{green!20}\textbf{0.02} \\ \hline
\multicolumn{1}{|c|}{0.50} & 0.89 & \multicolumn{1}{c|}{0.08} & 0.91 & \multicolumn{1}{c|}{\cellcolor{green!20}\textbf{-0.42}} & \cellcolor{green!20}\textbf{0.02} & \multicolumn{1}{c|}{0.06} & 0.9  & \multicolumn{1}{c|}{\cellcolor{green!20}\textbf{-0.44}} & \cellcolor{green!20}0.01 \\ \hline
\multicolumn{1}{|c|}{0.75} & 0.89 & \multicolumn{1}{c|}{0.08} & 0.91 & \multicolumn{1}{c|}{\cellcolor{green!20}\textbf{-0.67}} & \cellcolor{green!20}\textbf{0.02} & \multicolumn{1}{c|}{0.06} & 0.9  & \multicolumn{1}{c|}{\cellcolor{green!20}\textbf{-0.69}} & \cellcolor{green!20}0.01 \\ \hline
\rowcolor[HTML]{C0C0C0} 
\multicolumn{2}{|c|}{\cellcolor[HTML]{C0C0C0}B-VAE}            & \multicolumn{2}{c|}{\cellcolor[HTML]{C0C0C0}OPAL}      & \multicolumn{2}{c|}{\cellcolor[HTML]{C0C0C0}OPAL vs B-VAE}                       & \multicolumn{2}{c|}{\cellcolor[HTML]{C0C0C0}OPAL-AL}   & \multicolumn{2}{c|}{\cellcolor[HTML]{C0C0C0}OPAL-AL vs B-VAE}                  \\ \hline
\rowcolor[HTML]{C0C0C0} 
\multicolumn{1}{|c|}{\cellcolor[HTML]{C0C0C0}ME}      & Acc  & \multicolumn{1}{c|}{\cellcolor[HTML]{C0C0C0}ME} & Acc  & \multicolumn{1}{c|}{\cellcolor[HTML]{C0C0C0}$\Delta_{ME}$}    & $\Delta_{Acc}$      & \multicolumn{1}{c|}{\cellcolor[HTML]{C0C0C0}ME} & Acc  & \multicolumn{1}{c|}{\cellcolor[HTML]{C0C0C0}$\Delta_{ME}$}   & $\Delta_{Acc}$     \\ \hline
\multicolumn{1}{|c|}{0.25} & 0.68 & \multicolumn{1}{c|}{0.05} & 0.91 & \multicolumn{1}{c|}{\cellcolor{green!20}\textbf{-0.2}} & \cellcolor{green!20}\textbf{0.23} & \multicolumn{1}{c|}{0.05} & 0.9 & \multicolumn{1}{c|}{\cellcolor{green!20}\textbf{-0.2}} & \cellcolor{green!20}\textbf{0.22} \\ \hline
\multicolumn{1}{|c|}{0.50} & 0.68 & \multicolumn{1}{c|}{0.05} & 0.91 & \multicolumn{1}{c|}{\cellcolor{green!20}\textbf{-0.45}} & \cellcolor{green!20}\textbf{0.23} & \multicolumn{1}{c|}{0.05} & 0.9 & \multicolumn{1}{c|}{\cellcolor{green!20}\textbf{-0.45}} & \cellcolor{green!20}\textbf{0.22} \\ \hline
\multicolumn{1}{|c|}{0.75} & 0.68 & \multicolumn{1}{c|}{0.05} & 0.91 & \multicolumn{1}{c|}{\cellcolor{green!20}\textbf{-0.7}} & \cellcolor{green!20}\textbf{0.23} & \multicolumn{1}{c|}{0.05} & 0.9 & \multicolumn{1}{c|}{\cellcolor{green!20}\textbf{-0.7}} & \cellcolor{green!20}\textbf{0.22} \\ \hline
\rowcolor[HTML]{C0C0C0} 
\multicolumn{1}{|c|}{\cellcolor[HTML]{C0C0C0}Average} &      & \multicolumn{1}{c|}{\cellcolor[HTML]{C0C0C0}}   &      & \multicolumn{1}{c|}{\cellcolor[HTML]{C0C0C0}\cellcolor{green!20}\textbf{-0.315 (L)}} & \cellcolor{green!20}\textbf{0.0625 (L)} & \multicolumn{1}{c|}{\cellcolor[HTML]{C0C0C0}}   &      & \multicolumn{1}{c|}{\cellcolor[HTML]{C0C0C0}\cellcolor{green!20}\textbf{-0.37 (L)}} & \cellcolor{green!20}\textbf{0.052 (M)} \\ \hline
\end{tabular}

}
\end{table*}

\textbf{RQ3 Results.}
Tables~\ref{table:rq3_baselines_synthetic1}, \ref{table:rq3_baselines_synthetic2} and \ref{table:rq3_baselines_industry} show the average accuracy (Acc) and average manual effort (ME) achieved by \approach, \approach-AL and the four baselines across our three test-input validation datasets: \textsc{Synthetic-Pub1}, \textsc{Synthetic-Pub2}, and \textsc{Industry}. For each of the four baselines -- B-VAE, B-VIF, B-HiL-TV1, and B-HiL-TV2 -- the results are presented for three human effort levels: $25$\%, $50$\%, and $75$\%. For each human effort level, we report the differences in accuracy ($\Delta_{Acc}$) and manual effort ($\Delta_{ME}$) between \approach\ and each baseline, as well as between \approach-AL\ and each baseline.

A positive $\Delta_{Acc}$ means \approach\ (or \approach-AL) outperforms the baseline. These values are highlighted green when \approach's accuracy (or \approach-AL's accuracy) is equal to or higher than the baseline.  Similarly, $\Delta_{ME}$ shows the difference in manual effort between \approach\ (or \approach-AL) and the baseline. A negative $\Delta_{ME}$ means \approach\  (or \approach-AL)  outperforms the baseline and requires less effort. These values are highlighted green when \approach's effort   (or \approach-AL's effort) is equal to or lower than the baseline. The last rows of Tables~\ref{table:rq3_baselines_synthetic1}, \ref{table:rq3_baselines_synthetic2} and \ref{table:rq3_baselines_industry}   show the average $\Delta_{Acc}$ and $\Delta_{ME}$ for our three test-input validation datasets. 
We conducted Wilcoxon rank–sum test and Vargha-Delaney $\hat{A}_{12}$ test to compare accuracy and manual effort between the baselines and both OPAL and OPAL-AL. For comparisons yielding a significance level of \(p < 0.01\), the corresponding values of \(\Delta_{\mathrm{ME}}\) and \(\Delta_{\mathrm{Acc}}\) are emphasized in \textbf{boldface} in Tables~\ref{table:rq3_baselines_synthetic1}, \ref{table:rq3_baselines_synthetic2} and \ref{table:rq3_baselines_industry}. 
Furthermore, the comparison results show a large effect size for manual effort reduction across all datasets when comparing \approach\ and \approach-AL\ with baselines, and medium to large effect sizes for accuracy improvements across all datasets when comparing \approach\ and \approach-AL\ with baselines.

As shown in Tables~\ref{table:rq3_baselines_synthetic1}, \ref{table:rq3_baselines_synthetic2} and \ref{table:rq3_baselines_industry}, \approach\ reduces manual effort by 31.0\%, 24.0\% and 31.5\% and improves accuracy by 4.8\%, 2.5\% and 6.3\% on the Synthetic-Pub1, Synthetic-Pub2 and Industry datasets, respectively. With active learning, \approach-AL\ reduces manual effort by 35.2\%, 28.0\% and 37.0\% while improving accuracy by 4.7\%, 2.0\% and 5.2\% on the same datasets. Thus, adding the active-learning loop yields a further reduction in manual effort of 4.2\%, 4.0\% and 5.5\% across the three datasets.

In 32 out of 36 cases, \approach\ requires less manual effort than the baselines, while in 35 out of 36 cases, \approach's accuracy is equal to or higher than the baselines. 
In all cases (36 out of 36), \approach-AL requires less manual effort than the baselines, while in 30 out of 36 cases, \approach-AL's accuracy is equal to or higher than the baselines.

Table \ref{table:rq3_stat} reports the p-values from the Wilcoxon rank–sum test and the Vargha–Delaney $\hat{A}{12}$ effect sizes for validation accuracy and manual effort when comparing \approach\ against \approach-AL. Overall, on average,  \approach-AL requires $4.5$\%  less effort compared to \approach.  We find that neither approach significantly outperforms the other in terms of validation accuracy. In contrast, \approach-AL, significantly outperforms \approach\ in terms of manual effort with medium and large effect sizes. This shows that active learning loop can be considered as another factor to reduce manual effort. 

\begin{table}[t]
\centering
\caption{Wilcoxon rank-sum and Vargha-Delaney $\hat{A}_{12}$ statistical tests between \approach\ vs \approach\ extended with active learning (i.e., \approach-AL).
For comparisons yielding a significance level of \(p < 0.01\), the corresponding values of $\hat{A}_{12}$ are emphasized in \textbf{boldface}}
\label{table:rq3_stat}
\scalebox{.9}{
\begin{tabular}{|c|c|c|c|c|}
\hline
\textbf{Dataset} & \textbf{P-value (Acc)} & \textbf{$\hat{A}_{12}$ (Acc)} & \textbf{P-value (ME)} & \textbf{$\hat{A}_{12}$ (ME)} \\ \hline
\textbf{Synthetic-Pub1}     & 0.013            & \textbf{0.37 (M)}& 0.007            & \textbf{0.65(M)}\\ \hline
\textbf{Synthetic-Pub2}    & \textbf{$2.863\times10^{-15}$} & \textbf{0.48(S)} & \textbf{$9.051\times10^{-16}$} & \textbf{0.96(L)}\\ \hline
\textbf{Industry} & 0.041            & 0.43 (M)& 0.042            & 0.62(M)\\ \hline
\end{tabular}
}
\end{table}

\begin{tcolorbox}[breakable, colback=gray!10!white,colframe=black!75!black]
\textbf{Finding:} Across all the three test-input validation datasets, on average,  \approach\ requires 28.8\% less manual effort while achieving 4.5\% higher accuracy compared to the four test-input validation baselines. Similarly, \approach\ integrated with active learning, on average achieves 4\% higher accuracy while reducing the manual effort 33.4\% compared to the four test-input validation baselines. 

\textbf{Takeway:} Extending \approach\ with active learning improves its ability to further optimize the test-input validation effort by requiring an average of 4.5\% less manual labelling effort compared to not using an active learning loop.
\end{tcolorbox}

\section{Threats to Validity}

\emph{Internal validity.} 
In our experiments, we ensure that \approach\ and baselines are not fine-tuned on images that they will later label. Similarly, we prevent the MILP solver from processing images used during fine-tuning -- which might otherwise yield artificially high confidence levels. Specifically, we address these concerns by maintaining separate, non-overlapping \emph{fine-tuning}, \emph{optimization}, and \emph{to-be-labelled} subsets for each dataset throughout all experiments.
Recognizing that potential biases may arise from dataset characteristics, we note that four of the nine datasets studied in this article -- namely  \textsc{Cifar10}, \textsc{FashionMNIST}, \textsc{MNIST} and \textsc{SVHN} -- are public benchmarks that are widely used by the research community.  For these datasets, we directly used their existing labels as ground truth. The ground-truth labels for \textsc{Synthetic-Pub1} and \textsc{Synthetic-Pub2} were adopted from Hu et al.~\cite{marsha}. The \textsc{Industry} dataset was independently labelled by two human annotators (who are not co-authors), with the final labels reviewed and verified by domain experts, as detailed in Section~\ref{subsec:datasets}.

To address the randomness associated with data splitting and the fine-tuning of classifiers in RQ1, RQ2 and RQ3, each experimental configuration was repeated five times. To avoid potential data leakage, we note that vision models allow full control over which pre-trained weights to use. In RQ1 and RQ2, the pre-trained weights for VGG, ResNet, and ViT -- whether used in OPAL or the baselines -- are exclusively based on ImageNet. None of the nine datasets studied overlap with ImageNet, eliminating potential data contamination. Therefore, OPAL and the baselines are not fine-tuned on images that they later label.

In RQ2, for a fair comparison of \approach\ with the automated data-labelling baselines, \approach\ and the baselines were evaluated using an equal amount of manual effort to investigate each method's labelling accuracy. Similarly, in RQ3, fairness is ensured by comparing \approach, with and without active learning, to baselines using the same accuracy targets as those achieved by the baselines.

\emph{External validity.}
Our experiments use nine diverse datasets and three state-of-the-art deep learning  image classifiers. These datasets encompass a wide variety of images, including well-known benchmarks, real-world images, synthetic images, images of common-objects and industrial images. The three classifiers are based on deep learning architectures that have been extensively used across both research and industry~\cite{touvron_2021_training, wang_2016_temporal, girshick_2015_fastrcnn, boesch_2023_resnet_industry, parkhi_2015_vgg_industry, radford_2021_vit_industry}. The consistent results observed across different   datasets provide strong evidence of our findings’ generalizability. To further validate  generalizability, our study can be  replicated using other classifiers, such as DenseNet~\cite{densenet} and AlexNet~\cite{alexnet}, among others.

As with many widely used public datasets, the ground-truth labels used in our evaluation may contain noise. We did not manually validate all samples and instead relied on the original labels, consistent with common practice in prior work and across all compared baselines. Since all methods are evaluated on the same datasets with identical labels, any potential label noise affects all approaches equally and does not bias the comparative results. To assess the practical impact of this issue, we manually inspected the instances misclassified by \approach\ at the end of the RQ1 and observed that a non-trivial portion of these cases correspond to ambiguous samples or plausible labelling errors in the original datasets rather than clear failures of \approach.
\section{Discussion}

\textbf{\approach\ in the Software Development Lifecycle.} 
OPAL requires pre-trained classifiers as input; however, these classifiers are not assumed to be accurate or production-ready. Rather, they may be early prototypes, weak baseline models, or surrogate classifiers -- artifacts that commonly exist even in early stages of development~\cite{ng2018machine}. As a result, there is no conflict between the assumption of requiring pre-trained classifiers and the use of OPAL within the software development lifecycle. Testing activities -- such as test-input validation and oracle construction, which are the main tasks automated by OPAL -- often begin before a high-quality target classifier exists.  OPAL can support this setting: when classifiers are uncertain or disagree -- conditions typical of difficult problems -- OPAL automatically routes such cases to human annotators. This adaptive behaviour makes OPAL applicable to early-stage  testing scenarios, where automation must be used cautiously and manual effort must be allocated strategically.

\textbf{\approach\ versus Test Adequacy.}
The accuracy constraint in \approach\ should not be interpreted as a test-adequacy criterion. Adequacy criteria are used in the context of \emph{test generation}, i.e., generating a test suite that is ``adequate'' under some adequacy notion such as coverage for a given system under test (SUT). Our work, however, is not a test-generation method and does not propose an adequacy criterion for assessing whether a test suite is sufficient for revealing faults in a specific SUT. Instead, OPAL addresses a different (and complementary) testing problem: given an existing set of test inputs, OPAL optimizes the human effort required to (i) assign reliable test oracles (labels), and (ii) determine input validity (i.e., whether a generated or transformed input preserves the semantic content required by the underlying task such that it can be confidently labelled by a human expert). In this setting, the optimization objective is not ``how thoroughly does the suite test the SUT?'', but rather ``how reliably can we establish correct labels and validity judgments for the inputs we already have, under a user-specified accuracy target, with minimal manual effort?''. From this perspective, the accuracy constraint in OPAL refers to the correctness of the labelling and validation decisions themselves (oracle correctness and validity judgments), rather than to the accuracy of test results obtained from the SUT. Consequently, meeting OPAL’s label and validity accuracy target does not imply that a test suite is adequate for fault detection or that it satisfies any known test adequacy criteria. Rather, OPAL is complementary to established adequacy criteria: test-generation strategies guided by adequacy criteria can be used to construct a test suite for a given SUT, and OPAL can then be applied to that suite to ensure that the resulting test inputs have reliable oracles and validity labels before the suite is used for evaluation or verification.
\section{Related Work}
\label{sec:relwork}
Within the machine learning community, several supervised~\cite{vgg, resnet, vit} and semi-supervised~\cite{Lee_2013_PseudoLabel, Pham_2021_MetaPseudoLabel,Laine2017Temporal, Sajjadi2016Regularization, Berthelot2019MixMatch, Sohn_2021_fixmatch} methods have been proposed to effectively train classifiers on sparsely labelled datasets. Although these methods aim to maximize accuracy given an existing labelled dataset,  they neither allow control over the desired accuracy threshold nor provide a mechanism to minimize  labelling effort. As a result, they fail to use the available labelled data  in an optimal way to reach the highest attainable accuracy. In our evaluation, for instance, when provided with the same amount of labelled data that enables \approach\ to achieve an average accuracy of 98.8\%,  the baseline methods -- including both supervised~\cite{vgg, resnet, vit} and semi-supervised~\cite{Lee_2013_PseudoLabel,Pham_2021_MetaPseudoLabel} approaches -- cannot exceed 95.3\% average accuracy. Furthermore, supervised and semi-supervised learning often rely on a single classifier, whereas \approach\ uses multiple classifiers. While one could theoretically extend the baseline approaches to use multiple classifiers, any such extension would still need to address the problem of maximizing accuracy within a fixed labelling budget.   To our knowledge, no prior work has used these baselines to develop an accuracy-driven, effort-optimized solution for data labelling. In contrast, \approach\ employs MILP to achieve this goal both effectively and efficiently.

Parallel to supervised and semi-supervised methods, a growing body of work investigates using large language models (LLMs) directly as annotators. Ramanathan et al. propose an ``Automatic Data Labelling'' service that employs GPT models within an industrial labelling pipeline, using model-provided annotations optionally refined through human feedback and active learning \cite{ramanathan2025automatic}. Gilardi et al. show that ChatGPT can act as a high-quality zero-shot annotator for a wide range of text-annotation tasks, outperforming crowdworkers in accuracy \cite{gilardi2023chatgpt}. In computer vision, GPT-4V has been used for multi-attribute classification of person images \cite{fujimoto2024automatic}. At larger scales, MASSIVE illustrates how multilingual datasets can be automatically expanded via templating, translation, and annotation pipelines that incorporate LLM-based components \cite{fitzgerald2022massive}. 
These techniques generally treat the LLM as the primary annotator and evaluate its average annotation quality and cost. None of these LLM-based methods are explicitly accuracy-driven, i.e., aiming to achieve a target accuracy in their annotation task. In contrast, \approach\ formulates labelling as a constrained optimization problem, minimizing human effort subject to an explicit accuracy target, thus addressing a different goal and operational setting. 
As observed in recent studies, LLM-based labelling accuracy varies substantially across datasets and tasks (ranging approximately from 48\% to 94\%), indicating that LLMs do not yet achieve near-perfect or human-level reliability in many labelling scenarios. This gap is critical in testing and validation contexts, where even small amounts of mislabelling can significantly distort evaluation results. Consequently, an optimization-based approach such as OPAL is valuable for guiding accuracy–effort trade-offs and for identifying instances that are more likely to require human review in order to meet a desired accuracy target.
While our evaluation focuses on vision-based DNNs as classifiers, \approach\ is not limited to this setting. In particular, LLM-based annotators can be integrated into \approach\ as classifiers, provided they satisfy its core assumptions -- namely, that they produce both a label and an associated confidence score for each classification.
Evaluating \approach\ with LLM-based classifiers is left to future work.

Approaches for testing vision systems include metamorphic testing, which modifies existing labelled images while preserving their original labels \cite{marsha,WhenWhyTest2022,Guo_2018_DLFuzz,DBLP:conf/aitest/ArcainiBBG20}, as well as prompt-based generative AI -- such as Stable Diffusion \cite{Esser_2024_Stable_Diffusion, brooksInstructPix2PixLearningFollow2023} -- which synthesizes images from textual prompts and heuristically derives labels from these prompts. However, both methods often produce images that either fail to preserve the intended labels or are unrecognizable to humans~\cite{Ghobari_2025_active_learning, marsha, WhenWhyTest2022, Dola_2021_DAIV,Riccio_2020_SystematicMapping}. Recent studies classify such images as invalid test inputs, defining them as images that are under-represented in the training set of a given DL model, and suggest using out-of-distribution detection methods to identify them~\cite{WhenWhyTest2022, Dola_2021_DAIV, Stocco_2020_selforacle}. Specifically, DAIV~\cite{Dola_2021_DAIV} trains a VAE and uses its loss function to discriminate valid and invalid test inputs. Self-Oracle~\cite{Stocco_2020_selforacle} uses auto-encoders and time-series-based anomaly detection to identify invalid test inputs for autonomous driving systems. Riccio and Tonella~\cite{WhenWhyTest2022} have shown that automated test-input validators based on DAIV and Self-Oracle can achieve up to a 78\% agreement rate with human validators. Hu et al.~\cite{marsha} propose to discriminate between valid and invalid test inputs using  a metric based on VIF~\cite{vif}.

Ghobari et al. \cite{Ghobari_2025_active_learning} introduce a test-input validation method based on active learning and using 13 image-comparison metrics, including VIF and the VAE loss function. These metrics were previously used to evaluate the effectiveness of image-to-image translators in bridging the gap between simulated and real-world images for testing autonomous driving systems~\cite{stocco,Amini_2024_Translators}. Their work finds that both the use of active learning and the combination of multiple image-comparison metrics result in improvement over the baselines that rely exclusively on VIF or the VAE loss function.
Yet,  none of the above techniques are specifically designed to target a desired level of accuracy while minimizing human labelling effort. Our evaluation  in Section~\ref{sec:eval} indicates that \approach\ outperforms all the existing test-input validation methods we are aware of, namely the work of Ghobari et al.~\cite{Ghobari_2025_active_learning}, Hu et al.~\cite{marsha} and Riccio and Tonella~\cite{WhenWhyTest2022}. 
\section{Conclusion}

We introduce \approach, an accuracy-driven, effort-optimized data labelling and validation method that aims to meet a specified accuracy target with the least possible manual labelling  effort. Through a mixed-integer linear programming formulation for optimization, \approach\ achieves an average accuracy of 98.8\% on nine diverse datasets while reducing manual labelling effort by more than half. \approach\ significantly outperforms  state-of-the-art  baselines for test-data labelling and test-input validation. Further, we show that extending \approach\ with active learning leads to an additional 4.5\% reduction in manual labelling effort, without negatively impacting accuracy.

Existing data labelling and validation methods lack explicit control over the tolerance level for mislabelling; this can undermine reliable quality assurance of AI-enabled systems. \approach\ addresses this limitation by introducing a risk-driven strategy that allows practitioners to specify their desired accuracy target, ensuring more reliable data labelling and validation based on the risk associated with incorrect labels or invalid data. This in turn helps ensure that labelling quality aligns with the specific application needs and analytical goals at hand. 
Future work includes evaluating \approach\ on more complex, large-scale datasets such as ImageNet.
Another direction is to integrate LLMs as classifiers within \approach\ and empirically study the resulting accuracy–effort trade-offs.
Finally, \approach\ could be extended to not only assign labels from existing dataset classes, but also to identify invalid or out-of-distribution images by assigning a reject/invalid class.

\section{Code and Data Availability(Supporting Data)} 
Our \textbf{replication package} including our code, the studied public datasets and the DL classifiers is available online~\cite{github}.

\section*{Compliance with Ethical Standards}
\sectopic{Conflict of Interest.} The authors declare that they have no conflicts of interest.

\sectopic{Funding.} This research was supported by the Natural Sciences and Engineering Research Council of Canada (NSERC) through the Discovery and Discovery Accelerator.

\sectopic{Ethical Approval.} This research did not involve human participants or animals; therefore, ethical approval was not required.

\sectopic{Informed Consent.} No personal data or identifiable information is reported in this research; informed consent is not applicable.

\sectopic{Author Contributions.} This article is part of the first author's PhD studies, who led the work, including its conceptualization, methodology, analysis, and writing. The final two authors supervised the research, contributed to its direction and design, and provided substantial input on the writing and interpretation of the results.

\bibliographystyle{spbasic}
\bibliography{bibliography}

\end{document}